%% file: main.tex
\documentclass[10pt,twocolumn,letterpaper]{article}

\usepackage{cvpr}      
\usepackage{xcolor}         
\usepackage{graphicx}
\usepackage{multirow}
\usepackage{bbding}
\usepackage{array}
\usepackage{colortbl}
\usepackage{enumitem}
\usepackage{amsmath}
\usepackage{tcolorbox}
\tcbuselibrary{breakable}


\definecolor{cvprblue}{rgb}{0.21,0.49,0.74}
\usepackage[pagebackref,breaklinks,colorlinks,allcolors=cvprblue]{hyperref}

\def\paperID{5963} 
\def\confName{CVPR}
\def\confYear{2025}

\title{Optimus-2 \includegraphics[width=0.8cm]{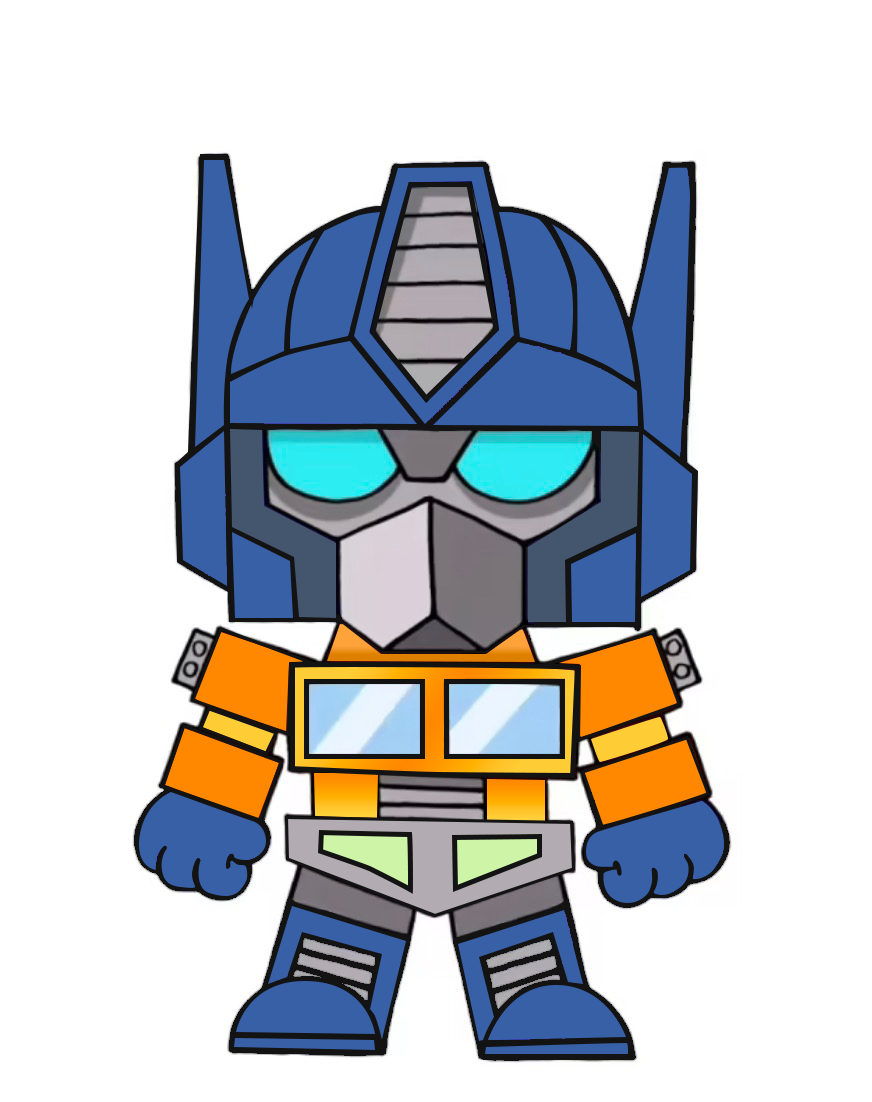}: Multimodal Minecraft Agent with Goal-Observation-Action Conditioned Policy}

\author{
Zaijing Li$^{1\,2}$,
Yuquan Xie$^{1}$,
Rui Shao$^{1}$\footnotemark[1], Gongwei Chen$^{1}$, Dongmei Jiang$^{2}$, Liqiang Nie$^{1}$\footnotemark[1]
\\ 
    $^1$Harbin Institute of Technology, Shenzhen 
  \hspace{10pt} $^2$Peng Cheng Laboratory \\
       \texttt{\{lzj14011,xieyuquan20016\}@gmail.com, 
       \{shaorui,nieliqiang\}@hit.edu.cn} \\
       \href{https://cybertronagent.github.io/Optimus-2.github.io/}{https://cybertronagent.github.io/Optimus-2.github.io/}
       }

\begin{document}
\maketitle
\renewcommand{\thefootnote}{\fnsymbol{footnote}} 
\footnotetext[1]{Corresponding authors}
\input{sec/0_abstract} 
\renewcommand{\thefootnote}{\arabic{footnote}}
\input{sec/1_intro}

\input{sec/2_related}
\input{sec/3_method}

\input{sec/4_exp}
\input{sec/5_conclusion}

\input{Supplementary/X_suppl}
\clearpage
{
    \small
    \bibliographystyle{ieeenat_fullname}
    \bibliography{main}
}


\end{document}

%% file: sec/0_abstract.tex
\begin{abstract}
Building an agent that can mimic human behavior patterns to accomplish various open-world tasks is a long-term goal. To enable agents to effectively learn behavioral patterns across diverse tasks, a key challenge lies in modeling the intricate relationships among observations, actions, and language. To this end, we propose Optimus-2, a novel Minecraft agent that incorporates a Multimodal Large Language Model (MLLM) for high-level planning, alongside a \textbf{G}oal-\textbf{O}bservation-\textbf{A}ction Conditioned \textbf{P}olicy \textbf{(GOAP)} for low-level control. GOAP contains (1) an Action-guided Behavior Encoder that models causal relationships between observations and actions at each timestep, then dynamically interacts with the historical observation-action sequence, consolidating it into fixed-length behavior tokens, and (2) an MLLM that aligns behavior tokens with open-ended language instructions to predict actions auto-regressively. Moreover, we introduce a high-quality \textbf{M}inecraft \textbf{G}oal-\textbf{O}bservation-\textbf{A}ction\textbf{ (MGOA)} dataset, which contains 25,000 videos across 8 atomic tasks, providing about 30M goal-observation-action pairs. The automated construction method, along with the MGOA dataset, can contribute to the community's efforts to train Minecraft agents. Extensive experimental results demonstrate that Optimus-2 exhibits superior performance across atomic tasks, long-horizon tasks, and open-ended instruction tasks in Minecraft. Please see the project page at \href{https://cybertronagent.github.io/Optimus-2.github.io/}{https://cybertronagent.github.io/Optimus-2.github.io/}.
\end{abstract}

%% file: sec/1_intro.tex
\section{Introduction}
\input{figures/fig-1}

Enabling agents to learn human behavioral patterns for completing complex tasks in open-world environments, is a long-standing goal in the field of artificial intelligence \cite{attribute-zhy,chen2024spa,shao2023detecting,li2023fine}. To effectively handle diverse tasks in an open-world environment like Minecraft \cite{qin2023mp5,li2024auto}, a prominent agent framework \cite{wang2023describe,wang2023jarvis,qin2023mp5,li2024optimus} integrates a task planner with a goal-conditioned policy. As illustrated in Figure \ref{fig:fig1} (left), this framework first utilizes the task planner’s language comprehension and visual perception abilities to decompose complex task instructions into sequential sub-goals. These sub-goals are then processed by a goal-conditioned policy to generate actions.

Although existing agents \cite{qin2023mp5,wang2023jarvis,li2024optimus} have made promising progress by using Multimodal Large Language Models (MLLM) \cite{chen2024lion,ye2024cat,shen2025mome} as planners, the current performance bottleneck for agents lies in the improvement of the goal-conditioned policy \cite{li2024optimus}. As the sub-goal serves as a natural language description of an observation-action sequence, the goal-conditioned policy needs to learn the crucial relationships among sub-goals, observations, and actions to predict actions. However, existing goal-conditioned policies exhibit the following limitations: \textbf{(1)} Existing policies neglect the modeling of the relationship between observations and actions. As shown in Figure \ref{fig:fig1}, they only model the relationship between the sub-goal and the current observation by adding the sub-goal embedding to the observation features \cite{lifshitz2024steve,cai2023groot,wang2024omnijarvis}. However, the current observation is generated by the previous action interacting with the environment. This implies a causal relationship between action and observation, which is neglected by current policies; \textbf{(2)} Existing policies struggle to model the relationship between open-ended sub-goals and observation-action sequences. As depicted in Figure \ref{fig:fig1}, existing policies primarily rely on either video encoders \cite{cai2023groot,wang2024omnijarvis} or conditional variational autoencoders (CVAE) \cite{lifshitz2024steve} as goal encoder to produce implicit goal embeddings. Such embeddings have limited representation ability \cite{wang2024omnijarvis}. Simply adding it to observation features is sub-optimal and unable to handle the complex relationship between sub-goals and observation-action sequences.

In this paper, we propose \textbf{Optimus-2}, a novel agent that incorporates an MLLM for planning, alongside a \textbf{G}oal-\textbf{O}bservation-\textbf{A}ction Conditioned \textbf{P}olicy (GOAP). To address the aforementioned challenges, we propose GOAP, which can better model the relationship among the observations, actions, and sub-goals in two aspects. 

\textbf{An Action-guided Behavior Encoder for observation-action sequence modeling.} To capture the relationship between observations and actions, the Action-guided Behavior Encoder first employs a Causal Perceiver to integrate action embeddings into observation features. It utilizes task-relevant action information as guidance to adjust the observation features, thereby providing fine-grained observation-action information for action prediction. Additionally, to model a long-term observation-action sequence without exceeding input length limitations, a History Aggregator is introduced to dynamically integrate current observation-action information with the historical sequence into fixed-length behavior tokens. Behavior tokens can capture the long-term dependencies of the observation-action sequence with a fixed and appropriate length. It enables the agent to predict actions that align with the logic of the observation-action sequence, rather than making isolated action predictions based solely on the current observation. 

\textbf{An MLLM to model the relationship between sub-goal and observation-action sequence.} To explicitly encode the semantics of sub-goals, we introduce an MLLM as the backbone of GOAP. It aligns the sub-goal with behavior tokens to predict subsequent actions auto-regressively. Leveraging the MLLM's language comprehension and multimodal perception capabilities, it can better integrate features from open-ended sub-goals and observation-action sequences, thereby enhancing the policy's action prediction ability. To the best of our knowledge, GOAP is the first effort to employ MLLM as the core architecture of a Minecraft policy, which demonstrates strong instruction comprehension capabilities for open-ended sub-goals.

Moreover, current Minecraft datasets either lack alignment among essential elements \cite{fan2022minedojo} or are not publicly accessible \cite{vpt}, resulting in a significant scarcity of high-quality observation-goal-action pairs necessary for policy training. To this end, we introduce an automated approach for constructing the \textbf{Minecraft Goal-Observation-Action (MGOA)} dataset. The MGOA dataset comprises 25,000 videos across 8 atomic tasks, providing approximately 30 million aligned observation-goal-action pairs. It will be made openly available to support advancements within the research community. We conducted comprehensive evaluations in the open-world environment of Minecraft, and the experimental results demonstrate that Optimus-2 achieves superior performance. Compared to previous SOTA, Optimus-2 achieves an average improvements of 27\%, 10\%, and 18\% on atomic tasks, long-horizon tasks, and open-ended sub-goal tasks, respectively.

In summary, our contributions are as follows:
\begin{itemize}
    \item We propose a novel agent Optimus-2, which consists of an MLLM for planning, and a policy for low-level control. The experimental results demonstrate that Optimus-2 exhibits superior performance on atomic tasks, long-horizon tasks, and open-ended sub-goal tasks.
    \item To better model the relationship among the observations, actions, and sub-goals, we propose Goal-Observation-Action Conditioned Policy, GOAP. It contains an Action-guided Behavior Encoder for observation-action sequence modeling, and an MLLM to model the relationship between sub-goal and observation-action sequence. 
    \item To address the scarcity of large-scale, high-quality datasets, we introduce the MGOA dataset. It comprises approximately 30 million aligned observation-goal-action pairs and is generated through an automated process without any manual annotations. The proposed dataset construction method and the released MGOA dataset can contribute to the community's efforts to train agents.
\end{itemize}

%% file: figures/fig-1.tex
\begin{figure}[htbp]
    \centering
    \includegraphics[width=0.5\textwidth]{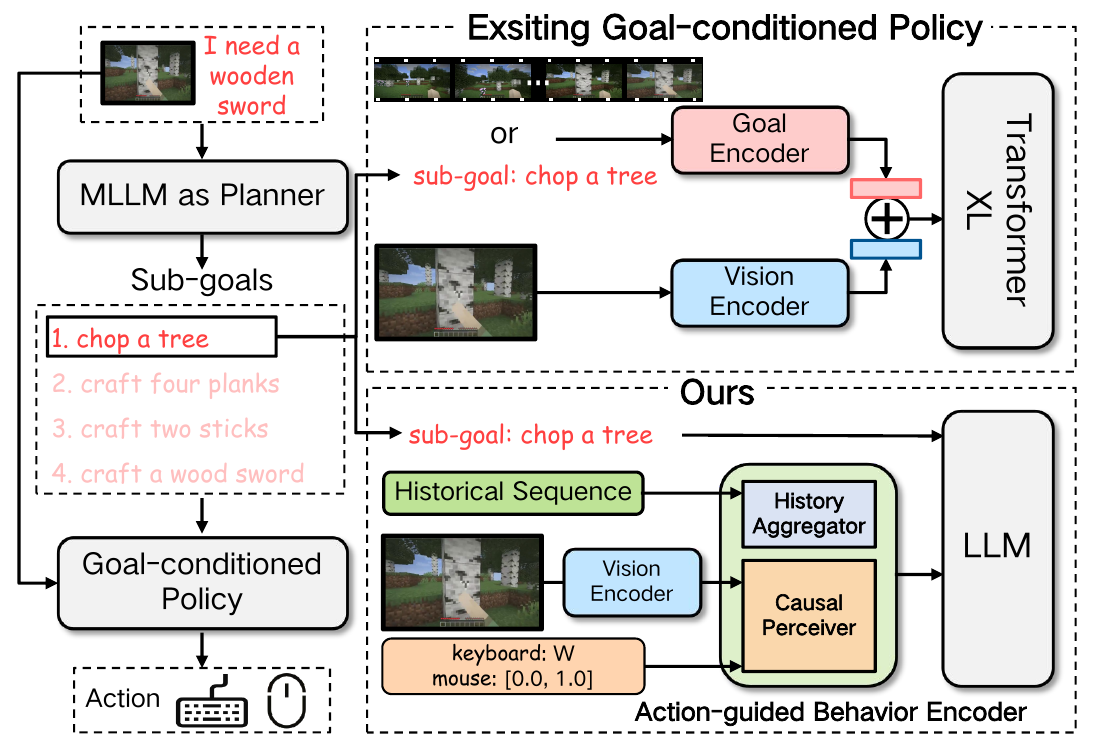}
    \caption{\textbf{Left}: General agent framework. \textbf{Right}: Comparison between existing goal-conditioned policies and ours. Existing Transformer-XL-based policies \cite{cai2023groot,lifshitz2024steve} exhibit limited natural language understanding capabilities and rely solely on combining implicit goal embeddings with visual embeddings as inputs. In contrast, our GOAP achieves superior action prediction by 1) employing an Action-guided behavior encoder to strengthen causal modeling between observations and actions, as well as to improve historical sequence modeling capabilities, and 2) leveraging MLLM to enhance open-ended language comprehension.}
    \label{fig:fig1}
\end{figure}


%% file: sec/2_related.tex
\section{Related Work}
\input{figures/fig-2}
\noindent\textbf{Minecraft Agents}. Previous works \cite{pmlr-v70-oh17a,ding2023clip4mc,cai2023open,hafner2023mastering} have constructed policies in Minecraft using reinforcement learning or imitation learning. VPT \cite{vpt} was training on large-scale video data recorded by human players, using behavior cloning to mimic human behavior patterns. GROOT \cite{cai2023groot} employs a video encoder as a goal encoder to learn semantic information from videos. However, these policies rely solely on visual observations as input and cannot follow human instructions to accomplish specific tasks. MineCLIP \cite{fan2022minedojo} introduces a video-text contrastive learning module as a reward model for policy, and STEVE-1 \cite{lifshitz2024steve} builds on VPT \cite{vpt} by incorporating MineCLIP as goal encoder, enabling policy to follow natural language instructions. Despite these advancements, these policies are constrained by language understanding and reasoning capabilities. To address this, current agents \cite{wang2023voyager,qin2023mp5,wang2023jarvis,li2024auto,li2024optimus,wang2024omnijarvis} leverage MLLM's instruction following capabilities to decompose complex tasks into executable sub-goal sequences, which are then fed into a goal-conditioned policy \cite{lifshitz2024steve,cai2023groot} or formed as executable code \cite{liu2024rl,zhu2023ghost,liu2024odyssey,zhao2023see}. Despite significant progress, the performance of current policies remains constrained by their limited ability to understand sub-goals. In this paper, we aim to develop an MLLM-based goal-conditioned policy to enhance the policy’s comprehension of open-ended sub-goals, thereby improving overall performance.

\noindent\textbf{Long-term Video Modeling}. Previous work \cite{vpt,lifshitz2024steve,fan2022minedojo,cai2023groot} have segmented videos into multiple clips for training to alleviate the challenges posed by long-sequence video inputs. However, this approach prevents the agent from learning comprehensive behavior representations from the entire video. To handle long-term video sequences \cite{li2025lion,zhang24aj,zhang2024flash}, existing studies employ temporal pooling \cite{maaz2023video}, querying transformers \cite{zhang2023video,he2024ma}, or token merging \cite{zhang2024token,song2024moviechat,jin2024chat} to integrate long-sequence visual tokens. Inspired by previous works \cite{chen2020memory,lee2018memory,lee2021video,wu2022memvit}, we propose a Q-former \cite{li2023blip,dai2023instructblip} structure with a memory bank \cite{he2024ma}, enabling effective long-term sequence modeling through interactions with historical queries. Unlike existing methods that model only the observation sequence, we focus on multimodal learning \cite{shao2019multi,shao2022open,shao2024detecting}. Moreover, different from previous work \cite{he2024ma} that primarily compress video features into fixed-length tokens, our Action-guided Behavior Encoder dynamically interacts with the historical sequence at each timestep, producing behavior tokens corresponding to the observation-action sequence from the start to the current timestep.


%% file: figures/fig-2.tex
\begin{figure*}[htbp]
    \centering
    \includegraphics[width=1.0\textwidth]{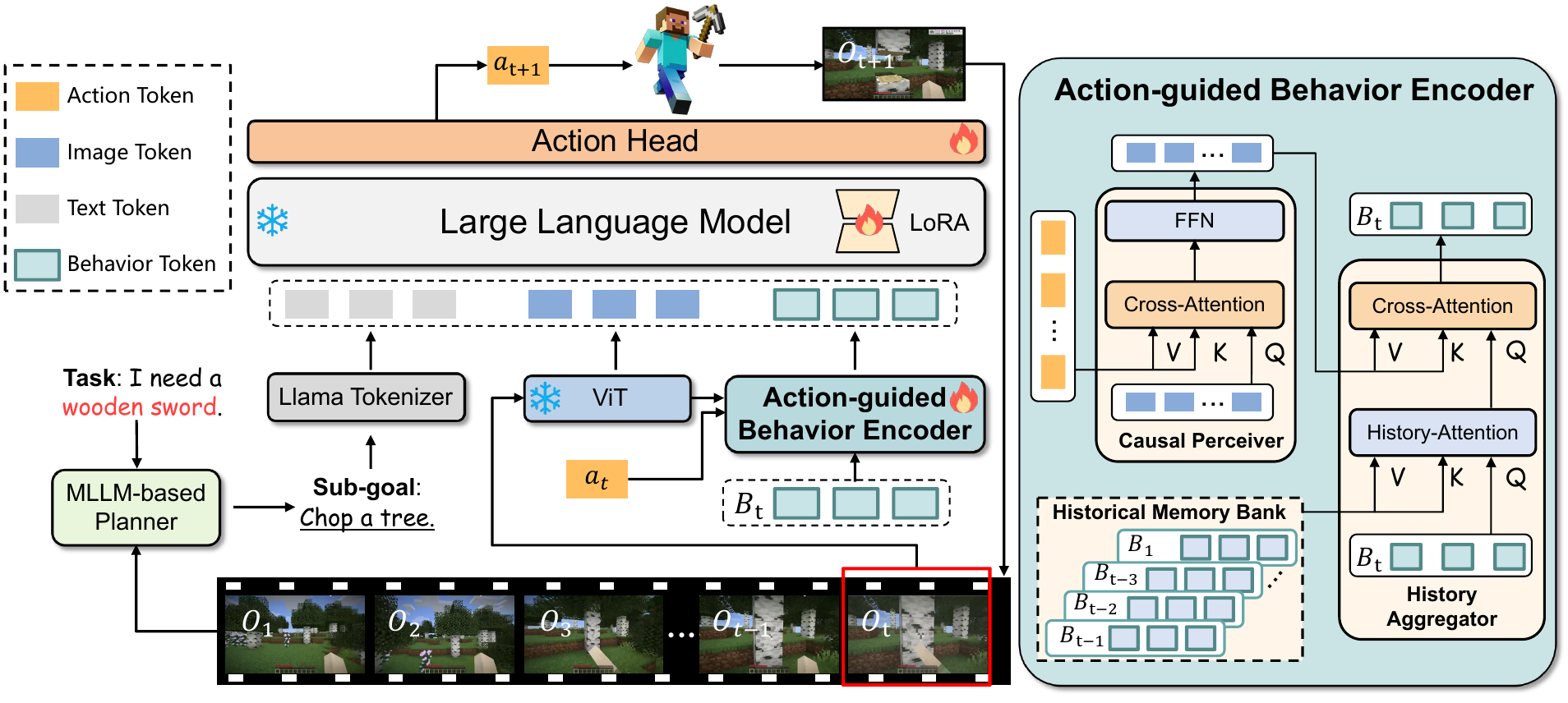}
    \caption{Overview of Optimus-2. Given a task and the current observation, Optimus-2 first uses an MLLM-based Planner to generate a series of sub-goals. Optimus-2 then sequentially executes these sub-goals through GOAP. GOAP obtains behavior tokens for the current timestep via the Action-guided Behavior Encoder, and these behavior tokens, along with image and text tokens, are fed into the LLM to predict subsequent actions.}
    \label{fig:fig2}
\end{figure*}

%% file: sec/3_method.tex
\section{Preliminaries and Problem Formulation}
In Minecraft, agents \cite{vpt,lifshitz2024steve,cai2023groot} exhibit behavior patterns similar to humans: at each time step \( t \), the agent receives a visual observation \( o_t \) and generates control actions \( a_{t+1} \) using the mouse and keyboard. These actions interact with the environment, resulting in a new visual observation \( o_{t+1} \). Through continuous interactions, a trajectory \( J =\) \(\{(o_1, a_1), (o_2, a_2), (o_3, a_3), \ldots, (o_T, a_T)\}\) is formed, where \( T \) represents the length of the trajectory. Previous work primarily trained Minecraft agents using reinforcement learning \cite{fan2022minedojo} or behavior cloning \cite{lifshitz2024steve,cai2023groot}. For example, in behavior cloning, the goal of the policy $p_{\theta}(a_{t+1}| o_{1:t}) $ is to minimize the negative log-likelihood of the actions at each time step \( t \) given the trajectory \( J \). Considering that such trajectories are typically generated under explicit or implicit goals, many recent approaches condition the behavior on a (implicit or explicit) goal \( g \) and learn goal-conditioned policy $p_{\theta}(a_{t+1}| o_{1:t}, g)$ \cite{lifshitz2024steve,cai2023groot}. Generally, for both agents and humans, the explicit goal \( g \) is a natural language instruction.

Formally, given a trajectory \( J \) with length \( T \), standard behavior cloning trains the policy $p_\theta(\cdot )$ with parameters $\theta$ by minimizing the negative log-likelihood of actions:

\begin{equation}
  \min_{\theta } \sum_{t=1}^{T} -\log_{}{p_\theta (a_{t+1} | o_{1:t},g)} 
\end{equation}

\section{Optimus-2}
In this section, we first give an overview of our proposed agent framework, Optimus-2. As shown in Figure \ref{fig:fig1} (left), it includes a planner for generating a series of executable sub-goals and a policy that sequentially executes these sub-goals to complete the task.

Next, we introduce how to implement Optimus-2's planner (Sec. \ref{planner}). Subsequently, we elaborate on how to implement the proposed GOAP (Sec. \ref{policy}). Finally, in Sec \ref{MGOA}, we introduce an automated dataset generation method to obtain a high-quality Minecraft Goal-Observation-Action dataset (MGOA) for training GOAP.


\subsection{MLLM-based Task Planner}
\label{planner}
In Minecraft, a complex task consists of multiple intermediate steps, i.e., sub-goals. For example, the task ``I need a wooden pickaxe'' includes five sub-goals: `chop a tree to get logs \includegraphics[width=0.3cm]{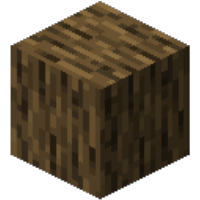}', `craft four planks \includegraphics[width=0.3cm]{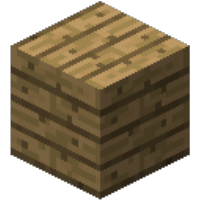}', `craft a crafting table \includegraphics[width=0.3cm]{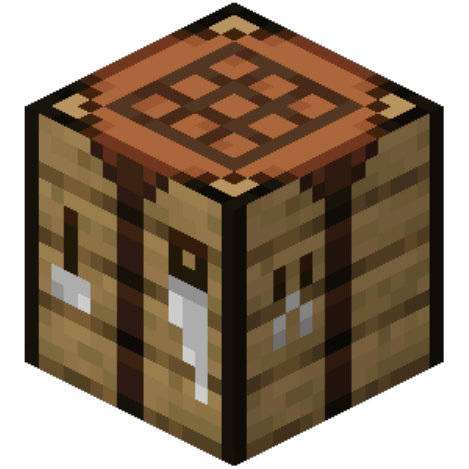}', `craft two sticks \includegraphics[width=0.3cm]{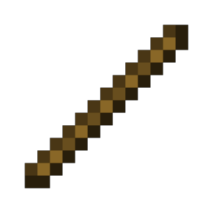}', and `craft a wooden pickaxe \includegraphics[width=0.3cm]{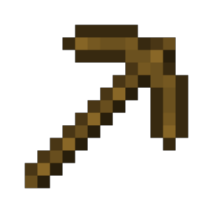}'. Therefore, a planner is essential for the agent, as it needs to decompose the given complex task into a sequence of executable sub-goals for the policy to execute sequentially. In this paper, we follow Li et al. \cite{li2024optimus}, employing an MLLM as the planner, which takes current observation and task instruction as input to generate sub-goals. 

\subsection{Goal-Observation-Action Conditioned Policy}
\label{policy}
According to Sec 3., a key insight into the relationship among observation $o$, action $a$, and sub-goal $g$ is: that the observation \( o \) and action \( a \) at the same time step have a causal relationship; and the sub-goal \( g \) is a natural language description of the observation-action sequence over a certain time. To better model the relationships among the three elements mentioned above, we propose first integrating the representations of observation and action at each time step, then modeling the observation-action sequences along the temporal dimension, and finally aligning the observation-action sequences with the sub-goal for action prediction.

Motivated by this, we propose a novel Goal-Observation-Action conditioned Policy, GOAP. As shown in Figure \ref{fig:fig2}, our GOAP consists of an Action-guided Behavior Encoder that dynamically models observation-action sequences into fixed-length behavior tokens and an MLLM that aligns such behavior tokens with sub-goal for action prediction.

\subsubsection{Action-guided Behavior Encoder}
Previous policies often overlook the causal relationship between observation and action at each timestep. Moreover, it remains a challenge to model the long-term observation-action sequence without exceeding input length constraints. To this end, we propose an Action-guided Behavior Encoder that integrates the representations of observation and action at each time step and then dynamically models the historical sequences into the fix-length behavior tokens. 
\input{table/tb_dataset}

Firstly, for the timestep $t$, we pass observation $o_t$ into a visual encoder $\texttt{VE}$ to obtain the visual features: 
\begin{equation}
 v_t \gets \texttt{VE}(o_t)
\end{equation}
where \( v_t \in \mathbb{R}^{P \times d} \), \( P \) is the number of patches for each image, and \( d \) is the dimension of the extracted image feature. In practice, we employ ViT \cite{dosovitskiy2020vit} as our visual encoder.

Then, we introduce a \textbf{Causal Perceiver} module to model the relationship between observations and actions. It takes the visual feature \( v_t \) as query tokens and the action embedding \( a_t \) as key and value. The module then constructs the information interaction between action \( a_t \) and \( v_t \) through a cross-attention mechanism:
\begin{equation}
\label{eq:kqv}
Q=v_t W_{v}^{Q}, K=a_t W_{a}^{K}, V=a_t W_{a}^{V}
\end{equation}
\begin{equation}
\label{eq:crossatt}
\hat{v}_{t}=CrossAttn(Q, K, V )=Softmax(\frac{QK^{T} }{\sqrt{d} } )V
\end{equation}
where \(W_{v}^{Q} \), \( W_{a}^{K} \), and \( W_{a}^{V} \) represent the weight matrices for the query (Q), key (K), and value (V), respectively. \(CrossAttn(\cdot)\) denotes the cross-attention layer, and \( d \) is the dimension of the image features. In this way, it explicitly assigns action information $a_{t}$ at time step $t$ to the visual features $\hat{v}_{t}$, enhancing the causal relationship between observations and actions.

Subsequently, we introduce a \textbf{History Aggregator} module to capture the information of the observation-action sequence along the temporal dimension, serving as the behavior representation. At each timestep $t$, behavior tokens $B_t$ serve as queries, while the sequence of historical behavior tokens $H_{t} = [B_{1}, B_{2}, \dots, B_{t-1}]$ acts as keys and values. The current behavior tokens interact with the historical sequence through a history-attention layer \(HisAttn(\cdot)\):
\begin{equation}
\hat{B}_{t}=HisAttn(Q, K, V )=Softmax(\frac{QK^{T} }{\sqrt{d} } )V
\end{equation}
where $Q$, $K$, and $V$ are calculated similarly to Eq \ref{eq:kqv}.

Finally, another cross-attention layer is introduced, using the behavior tokens $\hat{B}_{t}$ as queries, and the visual features $\hat{v}_{t}$ as keys and values. In this way, the behavior tokens incorporate the current observation-action information. Following the approach of He et al. \cite{he2024ma}, we construct a memory bank for historical behavior tokens \( H_{t} \), utilizing the similarity between adjacent features to aggregate and compress the behavior tokens. This method not only preserves early historical information but also keeps the historical behavior token sequence \( H_{t} \) at a fixed length to reduce computational costs. Leveraging the Action-guided Behavior Encoder, we obtain behavior tokens $\hat{B}_{t}$, which correspond to the observation-action sequence from the start to the current time step \( t \).


\subsubsection{MLLM Backbone}
To model the relationship between the sub-goal and observation-action sequence, we introduce an MLLM that takes the sub-goal $g$, current observation features $v_t$, and behavior tokens ${B}_{t}$ as input to predict subsequent actions auto-regressively. To enable the MLLM backbone \texttt{MLLM} to predict low-level actions, we employ VPT \cite{vpt} as action head \texttt{AH} to map output embeddings $\bar{a}_{t+1}$ of language model into the action space.
\begin{equation}
 \bar{a}_{t+1} \gets \texttt{MLLM}(\left [ g, v_t, B_t \right ] )
\end{equation}
\begin{equation}
 a_{t+1} \gets \texttt{AH}(\bar{a}_{t+1})
\end{equation}

Formally, given a dataset \( \mathcal{D} = \{(o_{1:T}, a_{1:T})\}_{M} \) with \( M \) complete trajectories, we train GOAP to learn the behavior distribution from $\mathcal{D}$ via behavioral cloning. Moreover, we introduce a KL-divergence loss to measure the output distribution similarity between GOAP and VPT \cite{vpt}. This helps our model effectively learn the knowledge from the teacher model VPT. The training loss can be formulated as follows:
\vspace{-5pt}
\begin{equation}
\begin{split}
\mathcal{L}_{\theta} = \lambda _{BC} \sum_{t=1}^{T} -\log_{}{p_\theta (a_{t+1} | o_{1:t},a_{1:t},g)} \\ 
+ \lambda _{KL} \sum_{t=1}^{T} D_{KL} ( q_\phi (a_{t+1} | o_{1:t}) \parallel p_\theta (a_{t+1} | o_{1:t},g)) 
\end{split}
\end{equation}
where $\lambda _{BC}$ and $\lambda _{KL}$ are trade off coefficients, $p_\theta$ is the GOAP, $q_\phi$ is the teacher model.

\input{table/tb_main_atomic}

\subsection{MGOA Dataset}
\label{MGOA}
In Minecraft, there remains a significant lack of high-quality goal-observation-action pairs to support behavior cloning training. Previous work has primarily relied on gameplay videos as training data. These datasets either lack natural language instructions (explicit goals) \cite{vpt,cai2023groot}, or use actions predicted by IDM models \cite{vpt} for each observation as pseudo-labels \cite{lifshitz2024steve,vpt}, which leads to a risk of misalignment between observations and actions. Inspired by Li et al. \cite{li2024optimus}, we propose an automated data generation pipeline that enables the creation of aligned goal-observation-action pairs without the need for manual annotations or human contractors. First, we utilize existing agents \cite{lifshitz2024steve}, providing them with clear natural language instructions to attempt task completion in Minecraft. We then record the actions and corresponding observations during goal execution, generating goal-observation-action pairs. 

\input{table/tb_main_horizon}

\input{table/tb_main_open}
To ensure the quality of the generated data, we apply the following filtering criteria: 1) only recording videos in which the task is successfully completed, and 2) discarding videos where task execution takes an excessive amount of time.  For more details, please refer to \textbf{Sup. C}. Through this automated approach, we obtained 25k high-quality Minecraft Goal-Observation-Action (MGOA) dataset. A comparison of the MGOA dataset with the existing Minecraft datasets is shown in Table \ref{tb:dataset}. Our automated data generation pipeline offers several key advantages: 1) it enables the generation of aligned goal-observation-action pairs without the need for manual annotation or pseudo-labeling; 2) its construction process is parallelizable, allowing for rapid dataset generation; and 3) it leverages local agents for data generation, resulting in low-cost production.

%% file: table/tb_dataset.tex
\begin{table}[tbp]
\centering
\caption{Comparison of the MGOA dataset with existing datasets. O, G, and A represent observation, goal, and action. VPT$^{\dag}$ indicates the amount of data that is openly accessible. MineCLIP$^{\ddagger}$ denotes narrated Minecraft videos available on YouTube.}
\label{tb:dataset}
\resizebox{0.5\textwidth}{!}{%
\begin{tabular}{llcccc}
\toprule[1.2pt]
 Format & Dataset  & O & G & A & \# Frames  \\ \hline
\multirow{2}{*}{Image-Text Pairs}   & MP5 \cite{qin2023mp5}  &   \Checkmark  & \Checkmark  &   &    500K       \\

  & OmniJARVIS \cite{wang2024omnijarvis}   & \Checkmark  & \Checkmark  & \Checkmark  &    600K      \\
\hline
\multirow{4}{*}{Gameplay Video}   & VPT$^{\dag}$ \cite{vpt}  &     \Checkmark &   & \Checkmark  &   6M   \\

 &   MineCLIP$^{\ddagger}$ \cite{fan2022minedojo}   & \Checkmark  & \Checkmark  &   &    20B      \\
 
&  STEVE-1 \cite{lifshitz2024steve}  & \Checkmark  & \Checkmark  & \Checkmark  &    32K   \\

\rowcolor[HTML]{E7EEFE}
    &   MGOA (Ours)  &  \Checkmark & \Checkmark  &  \Checkmark &  30M         \\ 
\bottomrule[1.2pt]
\end{tabular}
}
\end{table}

%% file: table/tb_main_atomic.tex

\begin{table}[t]
\centering
\caption{Main Result of GOAP on \textit{Atomic Tasks}. We report the average rewards of each task.}
\label{tb:main_atomic}
\vspace{-5pt}
\resizebox{0.5\textwidth}{!}{%

\begin{tabular}{lllll}
\toprule[1.2pt]
Policy    & Logs \includegraphics[width=0.3cm]{figures/logo/wood.pdf}  & Seeds \includegraphics[width=0.3cm]{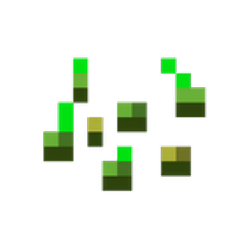}  & Dirt \includegraphics[width=0.3cm]{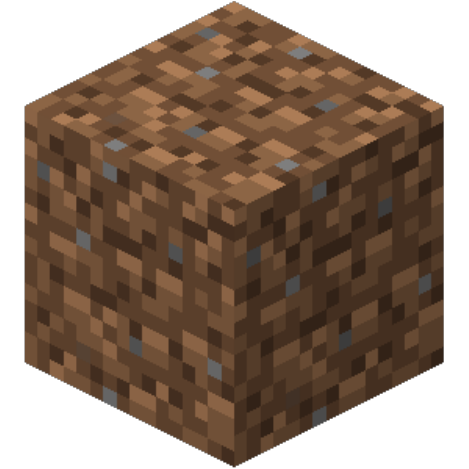} & Stone \includegraphics[width=0.3cm]{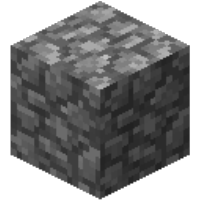}  \\ 
\hline
VPT (text) \cite{vpt}    &  2.6   &    0.8   &  9.2    &  0.0     \\
STEVE-1 \cite{lifshitz2024steve}    & 11.0   &  5.1  &   10.0   &    3.2   \\
GROOT  \cite{cai2023groot}     &  14.3  & 7.3   &   19.7   &    19.0   \\
FSQ GROOT \cite{wang2024omnijarvis}  & 10.8   &  8.2  &  20.3    &  5.8      \\
\rowcolor[HTML]{E7EEFE}
GOAP $_{[MLP]}$   &  7.2    & 4.3    & 14.4  & 15.5   \\ 
\rowcolor[HTML]{E7EEFE}
GOAP $_{[VPT]}$   &  \textbf{15.0}    & \textbf{8.5}    & \textbf{26.7}  & \textbf{25.7}   \\ 
\bottomrule[1.2pt]
\end{tabular}
}
\end{table}
\vspace{-5pt}

%% file: table/tb_main_horizon.tex

\begin{table*}[htbp]
\centering
\caption{Main Result of Optimus-2 on \textit{Long-horizon Tasks}. We report the average success rate (SR) on each task group, the results of each
task can be found in the \textbf{Sup. F.1}. Pure GPT-4V$^{\dag}$ denotes the use of GPT-4V in a zero-shot manner to generate executable sub-goals for the policy. Human$^{\ddagger}$ denotes the human-level baseline, with results sourced from previous work \cite{li2024optimus}.}
\label{tb:main_horizon}
\vspace{-5pt}
\renewcommand\arraystretch{1.1}
\begin{tabular}{llccccccc}
\toprule[1.2pt]
Method         & Policy         & Wood \includegraphics[width=0.3cm]{figures/logo/wood.pdf} & Stone \includegraphics[width=0.3cm]{figures/logo/cobblestone.pdf} & Iron \includegraphics[width=0.3cm]{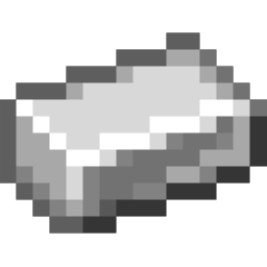} & Gold \includegraphics[width=0.3cm]{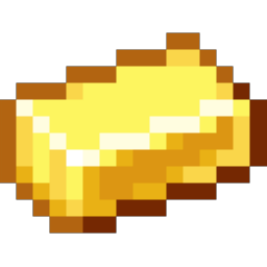} & Diamond \includegraphics[width=0.3cm]{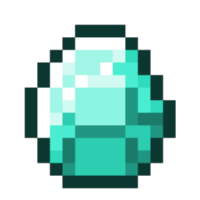} & RedStone \includegraphics[width=0.3cm]{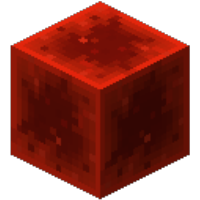} & Armor \includegraphics[width=0.3cm]{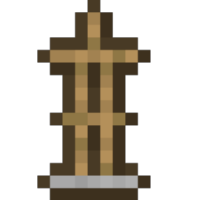}  \\ \hline
\hline
\multirow{3}{*}{Pure GPT-4V $^{\dag}$} &  VPT (text)     &    0.22  &   0.08    &   0.00  &   0.00   &     0.00    &       0.00   &    0.00           \\
 &  STEVE-1         &   0.41  &  0.20    &  0.00    & 0.00     &  0.00       &     0.00     &    0.00           \\
\rowcolor[HTML]{E7EEFE}
 \cellcolor{white} &  GOAP         &   \textbf{0.50}   &\textbf{ 0.31}      &   \textbf{0.12}   & \textbf{0.02}     &   \textbf{0.01}      &    \textbf{0.03}      &  \textbf{0.03}            \\ \hline
 \hline
DEPS \cite{wang2023describe} &  STEVE-1         &  0.77   &  0.48    &   0.16   &  0.00    &   0.01      &   0.00       &   0.10             \\
Jarvis-1 \cite{wang2023jarvis} & STEVE-1     &  0.93   &    0.89   &  0.36    &  0.07    &    0.08    &     0.16     &   0.15            \\
Optimus-1 \cite{li2024optimus} & STEVE-1    &  0.98   & 0.92      & 0.46    &   0.08   &     0.11    &   0.25       &   0.19           \\ 
\rowcolor[HTML]{E7EEFE}
Optimus-2 &  GOAP     & \textbf{0.99}     & \textbf{0.93}      & \textbf{0.53}     &   \textbf{0.09}  &    \textbf{0.13}     &     \textbf{0.28}     &   \textbf{0.21}      \\ \hline
\hline
\rowcolor[HTML]{FFF5DF}
Human$^{\ddagger}$ \cite{li2024optimus} &   \multicolumn{1}{c}{-}    & 1.00     &  1.00     &  0.86    &   0.17   &  0.16      &  0.33        &  0.28              \\ 

\bottomrule[1.2pt]
\end{tabular}
\vspace{-5pt}
\end{table*}


%% file: table/tb_main_open.tex
\begin{table}[htbp]
\centering
\caption{Main Result of GOAP on \textit{Open-Ended Instruction Tasks}. We report the average success rate (SR) on Torch \includegraphics[width=0.2cm]{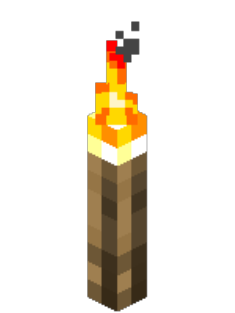}, Rail \includegraphics[width=0.3cm]{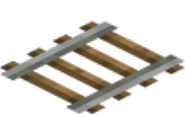}, Golden Shovel \includegraphics[width=0.3cm]{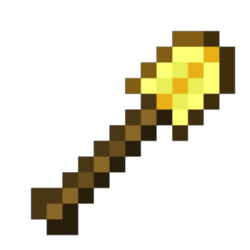}, Diamond Pickaxe \includegraphics[width=0.3cm]{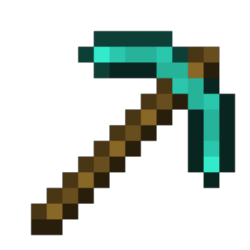}, and Compass \includegraphics[width=0.3cm]{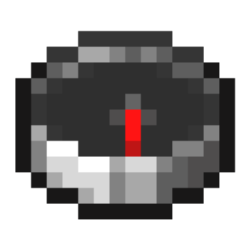}. GROOT \cite{cai2023groot} and FSQ GROOT \cite{wang2024omnijarvis} were not included as baselines, as they are unable to process language input.}
\label{tb:main_open}
\resizebox{0.5\textwidth}{!}{%
\begin{tabular}{llccccc}
\toprule[1.2pt]
Planner & Policy    & \includegraphics[width=0.2cm]{figures/logo/torch.pdf} & \includegraphics[width=0.3cm]{figures/logo/rail.pdf} & \includegraphics[width=0.3cm]{figures/logo/Golden_Shovel.pdf} & \includegraphics[width=0.3cm]{figures/logo/diamond_pickaxe.pdf} & \includegraphics[width=0.3cm]{figures/logo/compass.pdf} \\ \hline
\multirow{3}{*}{GLM-4V} & VPT (text)    &   0.05     & 0       &  0      &  0      & 0       \\
 & STEVE-1    &   0.60     &       0 &   0     &     0   &    0    \\
\rowcolor[HTML]{E7EEFE}
\cellcolor{white} & GOAP &    \textbf{0.71}    &    \textbf{0.39}    &    \textbf{0.11}    & \textbf{0.14}     &  \textbf{0.13}     \\  \hline
\multirow{3}{*}{GPT-4V} & VPT (text)    &  0.11      &  0      &     0   &     0   &  0      \\
 & STEVE-1    &  0.66      &   0.10     &    0    & 0       &  0      \\
\rowcolor[HTML]{E7EEFE}
\cellcolor{white} & GOAP &   \textbf{0.75}     & \textbf{0.47}       & \textbf{0.13}        &    \textbf{0.16}    &   \textbf{0.17}    \\
\bottomrule[1.2pt]
\end{tabular}
}
\vspace{-10pt}
\end{table}

%% file: sec/4_exp.tex
\section{Experiments}
\subsection{Experiments Setting}
\textbf{Environment}. Following \cite{vpt,lifshitz2024steve}, we conduct experiments in the complex, open-world environment of
Minecraft on the MineRL \cite{guss2019minerl} platform. The agent interacts with the MineRL environment at 20 frames per second, generating low-level control signals for the mouse and keyboard. For each task execution, the agent is initialized in a randomized environment, allowing us to evaluate the agent's generalization across diverse environments. Please refer to \textbf{Sup. B} for more details about the Minecraft environment. 

\noindent\textbf{Implementation details}. For the planner, we follow Li et al. \cite{li2024optimus}, using a hybrid multimodal memory empowered GPT-4V \footnote{https://openai.com/index/gpt-4v-system-card} as the agent’s planner. As for the policy, we initialize GOAP with the weights of DeepSeek-VL-1.3B \cite{lu2024deepseek} as initialization. We train it on the MGOA dataset and the publicly available OpenAI Contractor Dataset \cite{vpt} through behavior cloning. All experiments were conducted on 8x NVIDIA L40 GPUs. Training details and hyperparameter setting can be found in \textbf{Sup. D}.

\input{figures/fig-3}
\noindent\textbf{Evaluation Tasks \& Metrics}. 
Evaluation tasks are categorized into three types: \textit{Atomic Tasks}, \textit{Long-Horizon Tasks}, and \textit{Open-Ended Instruction Tasks}. For each task, the environment is randomly reinitialized on each attempt, with a minimum of 30 executions per task to ensure robustness.
\begin{itemize}
    \item \textit{Atomic Tasks} represent short-term skills in Minecraft. We select ``chop a tree to get logs \includegraphics[width=0.3cm]{figures/logo/wood.pdf}'', ``collect seeds \includegraphics[width=0.3cm]{figures/logo/seeds.pdf}'', ``collect dirt \includegraphics[width=0.3cm]{figures/logo/dirt.pdf}'', and ``mine stone \includegraphics[width=0.3cm]{figures/logo/cobblestone.pdf} with a pickaxe'' as evaluation tasks. These tasks evaluate the policy's basic capabilities in Minecraft. We report the average rewards (number of items obtained) per task execution as an evaluation metric.
    \item \textit{Long-horizon Tasks} consist of an interdependent atomic tasks sequence, where the failure of any single atomic task results in the failure of the entire sequence. These long-horizon tasks are designed to evaluate the agent’s capability to execute a series of diverse tasks continuously within a complex environment. We follow the setup of Li et al. \cite{li2024optimus}, conducting experiments on long-horizon tasks comprising 67 tasks grouped into 7 categories. We report the average Success Rate (SR) as an evaluation metric.
    \item \textit{Open-Ended Instruction Tasks} are not limited to predefined text formats; rather, they involve flexible language directives that prompt the agent to accomplish long-horizon tasks. These tasks evaluate the agent’s capacity to interpret and execute instructions expressed in open-ended natural language. We selected the Torch \includegraphics[width=0.25cm]{figures/logo/torch.pdf} from the Stone Group, Rail \includegraphics[width=0.3cm]{figures/logo/rail.pdf} from the Iron Group, Golden Shovel \includegraphics[width=0.3cm]{figures/logo/Golden_Shovel.pdf} from the Gold Group, Diamond Pickaxe \includegraphics[width=0.3cm]{figures/logo/diamond_pickaxe.pdf} from the Diamond Group, and Compass \includegraphics[width=0.3cm]{figures/logo/compass.pdf} from the Redstone Group as evaluation tasks. Given a crafting relationship graph, we instructed GPT-4V and GLM-4V \cite{glm2024chatglm} to generate five open-ended instructions for each task. This allows us to evaluate the policies' understanding and execution capabilities regarding open-ended instructions. Task instructions are provided in the \textbf{Sup. E.1}.
\end{itemize}

\input{table/tb_ablation}
\noindent\textbf{Baseline}. For \textit{Atomic Tasks} and \textit{Open-ended Instruction Tasks}, we compare GOAP with existing goal-conditioned policies, including VPT \cite{vpt}, STEVE-1 \cite{lifshitz2024steve}, GROOT \cite{cai2023groot} and FSQ GROOT \cite{wang2024omnijarvis}. For \textit{Long-horizon Tasks}, we employ GPT-4V, DEPS \cite{wang2023describe}, Jarvis-1 \cite{wang2023jarvis}, and Optimus-1 \cite{li2024optimus} as baselines. We also introduce a human-level baseline \cite{li2024optimus} to evaluate the performance gap between existing agents and human capabilities.

\subsection{Experimental Results}
The experimental results for Optimus-2 compared to the baselines across \textit{Atomic Tasks}, \textit{Long-horizon Tasks}, and \textit{Open-ended Instruction Tasks} are presented in Table \ref{tb:main_atomic}, Table \ref{tb:main_horizon}, and Table \ref{tb:main_open}, respectively. 

\textbf{GOAP excels in Atomic Tasks}. Table \ref{tb:main_atomic} shows that proposed GOAP achieves improvements of 5\%, 4\%, 31\%, and 35\% over the current SOTA on the Logs \includegraphics[width=0.3cm]{figures/logo/wood.pdf}, Seeds \includegraphics[width=0.3cm]{figures/logo/seeds.pdf}, Dirt \includegraphics[width=0.3cm]{figures/logo/dirt.pdf}, and Stone \includegraphics[width=0.3cm]{figures/logo/cobblestone.pdf}, respectively. These results demonstrate that GOAP has successfully mastered a range of short-term skills across diverse environments, and can acquire items more effectively than existing policies.


\textbf{Optimus-2 surpasses SOTA in Long-horizon Tasks}. Table \ref{tb:main_horizon} shows that Optimus-2 achieved the highest success rates across all seven task groups, particularly excelling in the challenging Diamond Group and Redstone Group with success rates of 13\% and 28\%, respectively. This indicates that Optimus-2 has effectively learned complex behavior patterns across atomic tasks, enabling it to sequentially execute multiple sub-goals and successfully complete long-horizon tasks within complex environments.

\input{figures/fig-6}
\textbf{GOAP outperforms in Open-ended Instruction Tasks}. As shown in Table \ref{tb:main_open}, GOAP achieved significantly higher success rates than existing agents across all tasks. Notably, on the challenging tasks of Golden Shovel \includegraphics[width=0.3cm]{figures/logo/Golden_Shovel.pdf}, Diamond Pickaxe \includegraphics[width=0.3cm]{figures/logo/diamond_pickaxe.pdf}, and Compass \includegraphics[width=0.3cm]{figures/logo/compass.pdf}, existing policies fail to complete these tasks, whereas GOAP achieves success rates of 13\%, 16\%, and 17\%, respectively. This advantage stems from GOAP’s superior comprehension of open-ended natural language instructions, whereas existing agents exhibit weaker instruction-following capabilities. Moreover, Figure \ref{fig:fig3} illustrates an example of different policies executing an open-ended goal. Due to the limited representation capability of their goal encoders, VPT \cite{vpt} and STEVE-1 \cite{lifshitz2024steve} fail to understand the goal, ``I need some iron ores, what should I do?'' In contrast, GOAP leverages the MLLM's understanding of open-ended instructions to effectively accomplish the goal (obtaining iron ore \includegraphics[width=0.3cm]{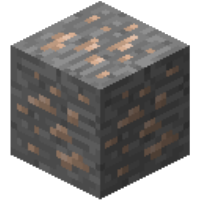}).




\subsection{Ablation Study}
There are many unexplored questions around best practices for developing MLLM-based policy in Minecraft. In this section, we conduct an extensive ablation study and summarize our key findings. 

\input{figures/fig-4}
\textbf{The Action-guided Behavior Encoder plays a crucial role in task execution}. As shown in Table \ref{tb:ablation}, the removal of the Causal Perceiver leads to an average performance decline of 42\% across all tasks, highlighting the importance of capturing the causal relationship between observations and actions. Moreover, eliminating the History Aggregator and Memory Bank also results in an average performance decline of 36\% across all tasks. This emphasizes the crucial role of the History Aggregator in modeling observation-action sequences and the Memory Bank in dynamically storing long-sequence information.

\textbf{LLM significantly enhances policy's ability to understand open-ended instructions}. As shown in Figure \ref{fig:fig6}, replacing the LLM backbone with a Transformer-XL leads to a noticeable decline in performance. We attribute this to the LLM's pretraining on large-scale textual corpora, which endows it with a robust comprehension of open-ended language, a capability that Transformer-XL lacks.

\textbf{A pretrained action head improves performance in Minecraft}. As shown in Table \ref{tb:main_atomic}, replacing VPT with a 2-layer MLP projector as the action head leads to a noticeable decline in Optimus-2’s performance. While MLP-based action heads have shown promising results in other domains \cite{kim2024openvla,liu2024rdt}, this substitution is less effective in the Minecraft environment. We attribute this to VPT’s extensive pretraining on large-scale gameplay data, which equips it with substantial domain-specific knowledge critical for effective task execution in Minecraft.


\textbf{The MGOA datsaset is beneficial for training GOAP}. We conducted comparative experiments to evaluate the impact of different training datasets on performance. As shown in Figure \ref{fig:fig4}, training only with the current most commonly used dataset, OpenAI Contractor Dataset (OCD), results in suboptimal performance for GOAP on all \textit{Atomic Tasks}. For example, compared to training with a mixed dataset, its performance on Stone \includegraphics[width=0.3cm]{figures/logo/cobblestone.pdf} dropped by 89\%. We attribute this to the fact that OCD offers a wide variety of tasks but lacks high data quality. In contrast, using our MGOA dataset, performance on the four atomic tasks improved by an average of 70\% compared to using only the OCD data. We attribute this to the fact that MGOA contains high-quality aligned goal-observation-action pairs, which is beneficial for policy training. Further, we mix the two datasets to train the policy in order to balance task diversity and data quality, leading to improved performance.

\input{figures/fig-5}
\subsection{Visualization of Behavior Representation}

As shown in Figure \ref{fig:fig5}, we apply t-SNE \cite{tsne} to visualize observation features extracted by ViT \cite{dosovitskiy2020vit}, MineCLIP \cite{fan2022minedojo}, and the Action-guided Behavior Encoder for four tasks. From (a) and (b) in Figure \ref{fig:fig5}, it is evident that the behavior representations extracted by ViT and MineCLIP are highly mixed, making it challenging to delineate the boundaries between different tasks. This lack of clear distinction between task-specific behavior representations can hinder the model's ability to understand the unique behavior patterns associated with each task, potentially leading to task failure. In contrast, the visualization in (c) of Figure \ref{fig:fig5} reveals clear, distinct clusters for each task, demonstrating that the Action-guided Behavior Encoder effectively captures subtle differences in observation-action sequences, thereby learning robust behavior representations across tasks.

%% file: figures/fig-3.tex
\begin{figure*}[htbp]
    \centering
    \includegraphics[width=1\textwidth]{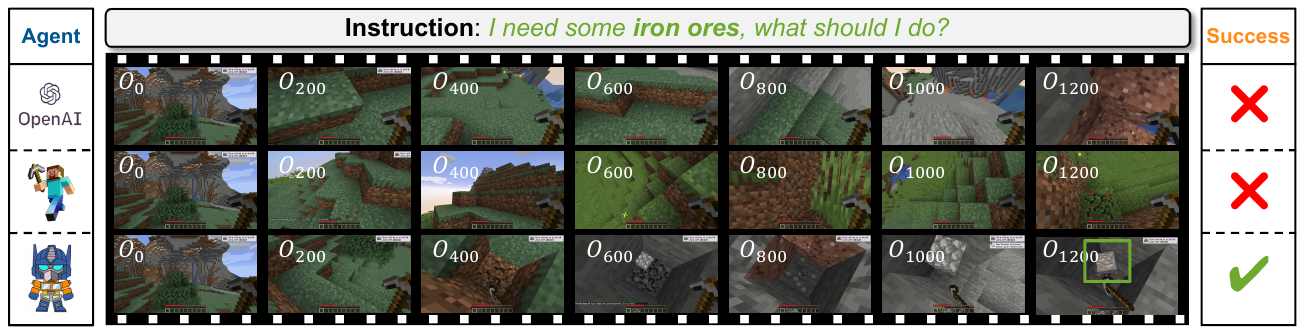}
    \caption{An illustration of VPT (text) \cite{vpt}, STEVE-1 \cite{lifshitz2024steve}, and Optimus-2 executing the open-ended instruction, ``I need some iron ores, what should I do?''. Existing policies are limited by their instruction comprehension abilities and thus fail to complete the task, whereas GOAP leverages the language understanding capabilities of the MLLM, enabling it to accomplish the task.}
    \label{fig:fig3}
    \vspace{-5pt}
\end{figure*}

%% file: table/tb_ablation.tex
\begin{table}[t]
\centering
\caption{Ablation study of Action-guided Behavior Encoder on \textit{Atomic Tasks}. We report average rewards on each task. \texttt{CP.}, \texttt{HA.}, and \texttt{MB.} represent the Causal Perceiver, History Aggregator, and Memory Bank, respectively.}
\label{tb:ablation}
\renewcommand\arraystretch{1.1}
\resizebox{0.5\textwidth}{!}{%
\begin{tabular}{ccc|ccccc}
\toprule[1.1pt]
\multicolumn{3}{c|}{Ablation Setting}                                                                         & \multicolumn{5}{c}{Atomic Task}          \\ \hline
\texttt{CP.}       & \texttt{HA.}                & \texttt{MB.}                                                & Logs \includegraphics[width=0.25cm]{figures/logo/wood.pdf}  & Seeds \includegraphics[width=0.25cm]{figures/logo/seeds.pdf} & Dirt \includegraphics[width=0.25cm]{figures/logo/dirt.pdf}  & Stone \includegraphics[width=0.25cm]{figures/logo/cobblestone.pdf} & Average  \\ \hline

\rowcolor[HTML]{E7EEFE}
\Checkmark & \Checkmark   &       \Checkmark & \textbf{15.0} & \textbf{8.5} &\textbf{26.7}  &\textbf{25.7} & \textbf{19.0} \\
   &  &  &  6.1    &  5.4    &    12.7 &      15.7  & 10.0 \textcolor{red}{\scriptsize ($\downarrow$ 47.4\%)}    \\
\Checkmark &    &           & 10.2 & 4.7  &  12.8  & 21.1  & 12.2 \textcolor{red}{\scriptsize ($\downarrow$ 35.8\%)}      \\
 &     \Checkmark     &   \Checkmark        &  7.4                           & 6.2  &   13.1    &   15.5  & 10.6 \textcolor{red}{\scriptsize ($\downarrow$ 44.2\%)}        \\
\Checkmark &  \Checkmark   &         &   12.0        & 6.8    & 22.7  & 16.8   & 14.6 \textcolor{red}{\scriptsize ($\downarrow$ 23.2\%)}    \\
\bottomrule[1.1pt]
\end{tabular}
}
\end{table}

%% file: figures/fig-6.tex
\begin{figure}[htbp]
    \centering
    \includegraphics[width=0.4\textwidth]{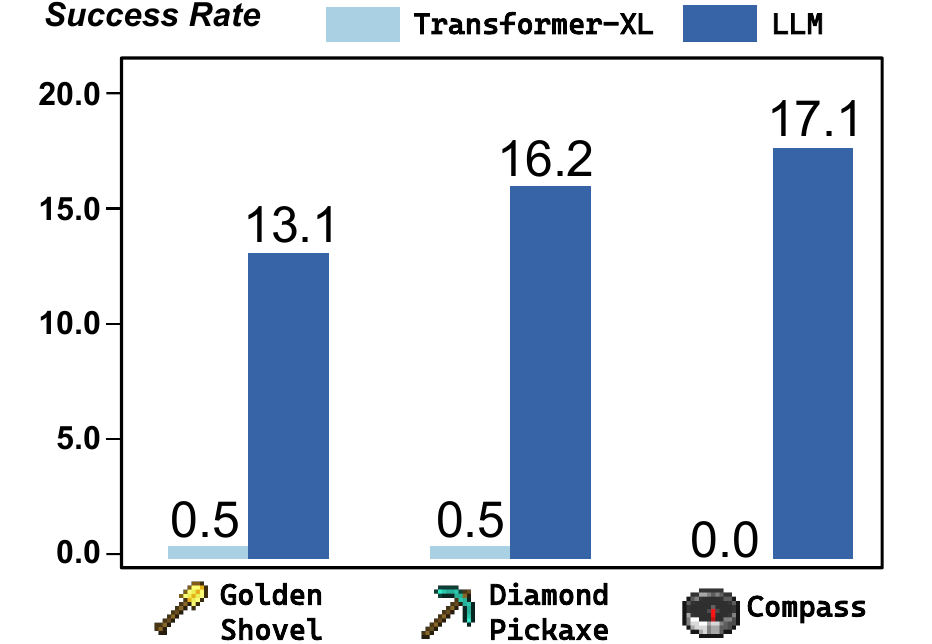}
    \caption{Ablation of LLM backbone on \textit{Open-ended Instruction Tasks}, Golden Shovel \includegraphics[width=0.3cm]{figures/logo/Golden_Shovel.pdf}, Diamond Pickaxe \includegraphics[width=0.3cm]{figures/logo/diamond_pickaxe.pdf}, and Compass \includegraphics[width=0.3cm]{figures/logo/compass.pdf}.}
    \label{fig:fig6}
    \vspace{-6pt}
    
\end{figure}

%% file: figures/fig-4.tex
\begin{figure}[htbp]
    \centering
    \includegraphics[width=0.4\textwidth]{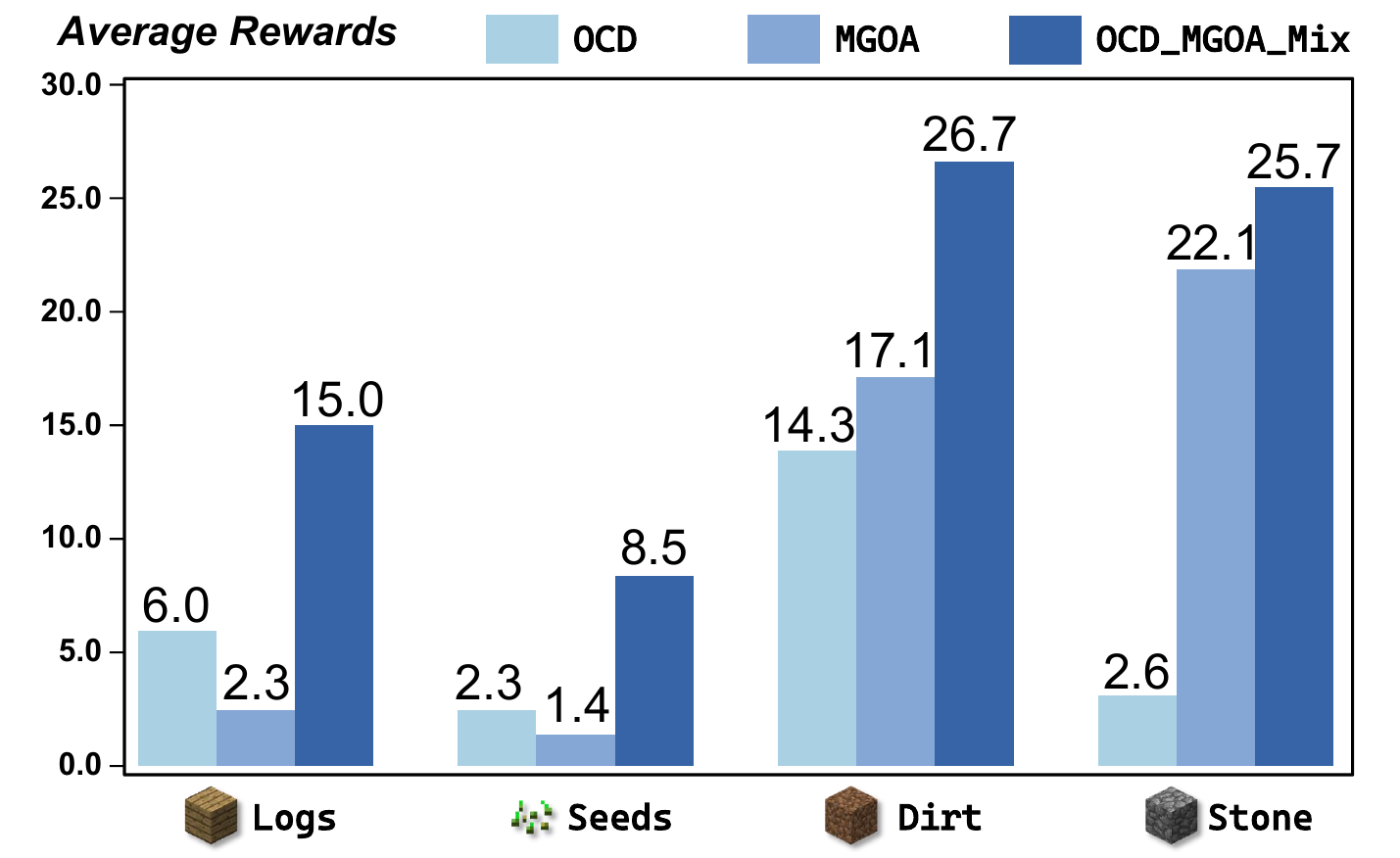}
    \caption{Ablation study on Training data. OCD refers to the OpenAI Contractor Dataset \cite{vpt}. We report the average rewards on each \textit{Atomic Task}.}
    \label{fig:fig4}
    \vspace{-6pt}
\end{figure}

%% file: figures/fig-5.tex
\begin{figure}[t]
    \centering
    \includegraphics[width=0.5\textwidth]{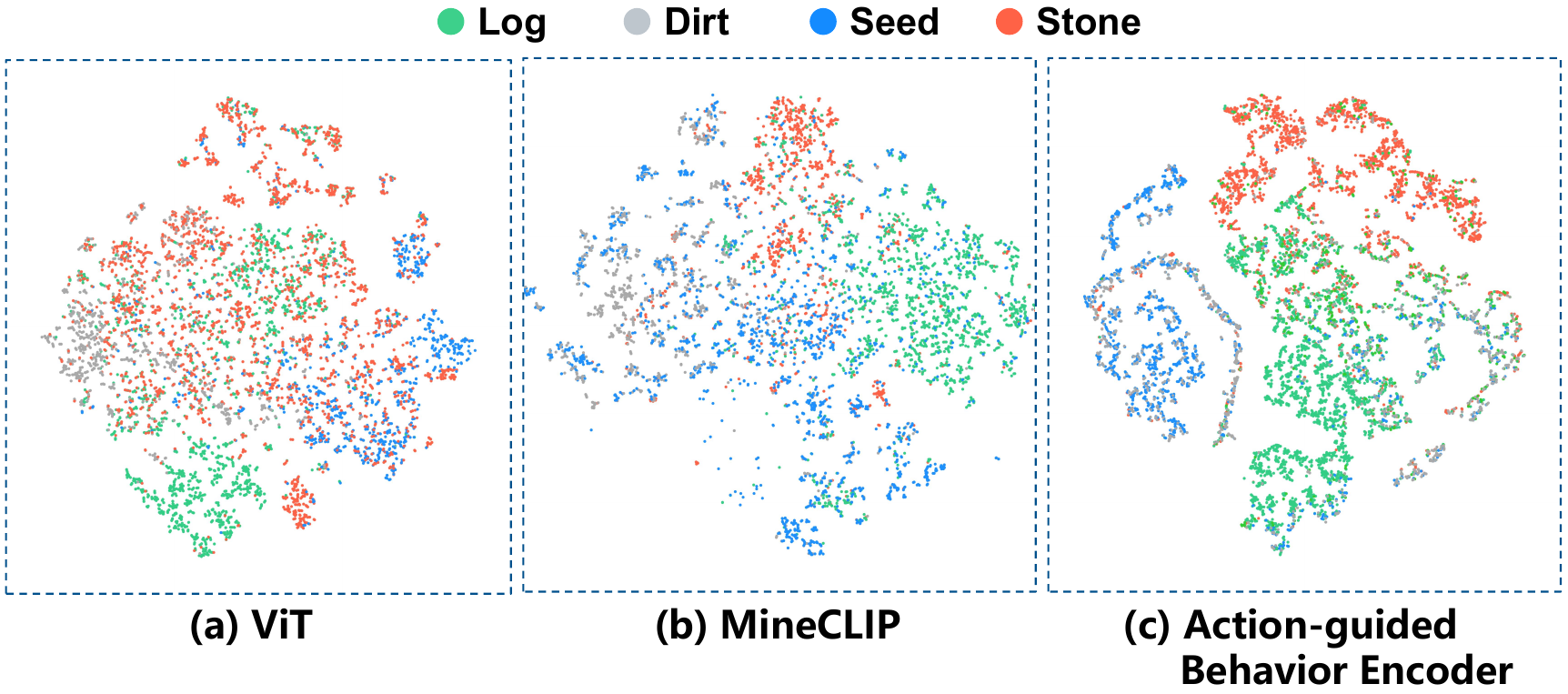}
    \caption{t-SNE visualization of representations extracted by (a) ViT (b) MineCLIP and (c) Action-guided Behavior Encoder across \textit{Atomic Tasks}. The visualization results show that the representations in (a) and (b) cannot distinguish between different tasks, whereas our Action-guided Behavior Encoder clearly differentiates the behavior representations for the four tasks.}
    \label{fig:fig5}
    \vspace{-6pt}
\end{figure}

%% file: sec/5_conclusion.tex
\section{Conclusion}
In this paper, we propose a novel agent, Optimus-2, which can excel in various tasks in the open-world environment of Minecraft. Optimus-2 integrates an MLLM for high-level planning and a Goal-Observation-Action conditioned Policy (GOAP) for low-level control. As a core contribution of this paper, GOAP includes an Action-guided Behavior Encoder to model the observation-action sequence and an MLLM to align the goal with the observation-action sequence for predicting subsequent actions. Extensive experimental results demonstrate that GOAP has mastered various atomic tasks and can comprehend open-ended language instructions. This enables Optimus-2 to achieve superior performance on long-horizon tasks, surpassing existing SOTA. Moreover, we introduce a Minecraft Goal-Observation-Action dataset to provide the community with large-scale, high-quality data for training Minecraft agents.

%% file: Supplementary/X_suppl.tex
\clearpage
\setcounter{page}{1}
\setcounter{section}{0}
\maketitlesupplementary
\renewcommand{\thesection}{\Alph{section}}

\noindent The supplementary document is organized as follows:

\begin{itemize}
\setlength{\itemsep}{10pt}
    \item Sec. \ref{sup:limitation}: Limitation and Future Work.
    
    \item Sec. \ref{sup:mc}: Minecraft Environment.
    
    \item Sec. \ref{sup:dataset}: MGOA Dataset.
    
    \item Sec. \ref{sup:train}: Training Details.
    
    \item Sec. \ref{sup:evaluate}: Evaluation Benchmark.
    
    \item Sec. \ref{sup:exp}: Experimental Results.

    \item Sec. \ref{sup:case}: Case Study.
\end{itemize}

\section{Limitation and Future Work}
\label{sup:limitation}
In this paper, we aim to explore how agents can mimic human behavior patterns in Minecraft to accomplish various tasks. Experimental results demonstrate that Optimus-2 performs exceptionally well in both atomic tasks and long-horizon tasks. However, due to the lack of sufficient high-quality data for open-ended tasks (such as ``building a house'' and ``defeating the Ender Dragon''), there remains significant room for improvement. Once such datasets are available, the ability of Optimus-2 to complete open-ended tasks will be enhanced. Moreover, despite showing promising performance in Minecraft, we have not yet extended our exploration to other simulation platforms, which represents a potential direction for future research.

\section{Minecraft}
\label{sup:mc}
\input{Supplementary/picture/behavior}
Minecraft is an extremely popular sandbox video game developed by Mojang Studios \footnote{https://www.minecraft.net/en-us/article/meet-mojang-studios}. It allows players to explore a blockly, procedurally generated 3D world with infinite terrain, discover and extract raw materials, craft tools and items, and build structures or earthworks. In this enviroment, AI agents need to face situations that are highly similar to the real world, making judgments and decisions to deal with various environments and problems. As shown in Figure \ref{fig:behavior}, both agents and humans are required to receive natural language instructions and current observations as input, and then output low-level actions, such as mouse and keyboard control commands. Therefore, Minecraft serves as an ideal open-world environment for training agent that can learn human behavior patterns.

\subsection{Basic Rules}
\noindent\textbf{Biomes.} The Minecraft world is divided into different areas called ``biomes''. Different biomes contain different blocks and plants and change how the land is shaped. There are 79 biomes in Minecraft 1.16.5, including ocean, plains, forest, desert, etc. Diverse environments have high requirements for the generalization of agents.

\noindent\textbf{Item.} In Minecraft 1.16.5, there are $975$ items can be obtained, such as wooden pickaxe \includegraphics[width=0.25cm]{figures/logo/wooden_pickaxe.pdf}, iron sword \includegraphics[width=0.25cm]{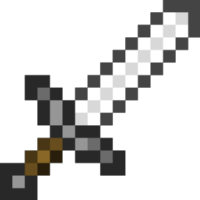}. Item can be obtained by crafting or destroying blocks or attacking entities. For example, agent can attack cows \includegraphics[width=0.25cm]{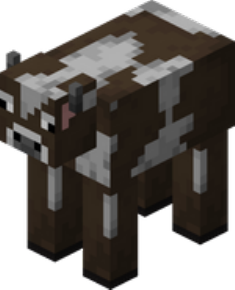}  to obtain leather \includegraphics[width=0.25cm]{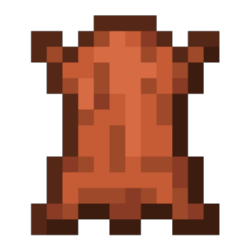} and beef \includegraphics[width=0.25cm]{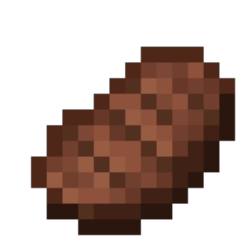}. Agent also can use $1$ stick \includegraphics[width=0.25cm]{figures/logo/stick.pdf} and $2$ diamonds \includegraphics[width=0.25cm]{figures/logo/diamond.pdf} to craft diamond sword \includegraphics[width=0.25cm]{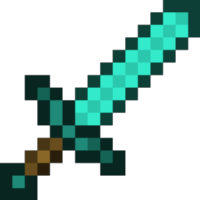}.

\noindent\textbf{Technology Tree.} In Minecraft, the technology hierarchy comprises six levels: wood \includegraphics[width=0.25cm]{figures/logo/wood.pdf}, stone \includegraphics[width=0.25cm]{figures/logo/cobblestone.pdf}, iron \includegraphics[width=0.25cm]{figures/logo/iron_ingot.pdf}, gold \includegraphics[width=0.25cm]{figures/logo/gold_ingot.pdf}, diamond \includegraphics[width=0.25cm]{figures/logo/diamond.pdf}, and redstone \includegraphics[width=0.25cm]{figures/logo/readstone.pdf}. Each tool level corresponds to specific mining capabilities. Wooden tools can mine stone-level blocks but are incapable of mining iron-level or higher-level blocks. Stone tools can mine iron-level blocks but cannot mine diamond-level or higher-level blocks. Iron tools are capable of mining diamond-level blocks. Finally, diamond tools can mine blocks of any level, including redstone-level.

\input{Supplementary/table/action_space}

\noindent\textbf{Gameplay progress.} Progression in Minecraft primarily involves discovering and utilizing various materials and resources, each unlocking new capabilities and opportunities. For instance, crafting a wooden pickaxe \includegraphics[width=0.25cm]{figures/logo/wooden_pickaxe.pdf} enables players to mine stone \includegraphics[width=0.25cm]{figures/logo/cobblestone.pdf}, which can then be used to create a stone pickaxe \includegraphics[width=0.25cm]{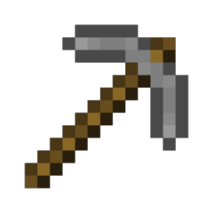} and a furnace \includegraphics[width=0.25cm]{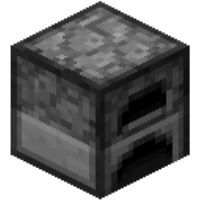}. These tools allow for the mining and smelting of iron ore \includegraphics[width=0.25cm]{figures/logo/iron_ore.pdf}. Subsequently, crafting an iron pickaxe \includegraphics[width=0.25cm]{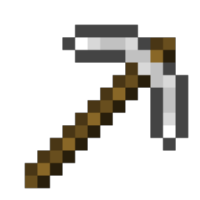} enables the extraction of diamonds \includegraphics[width=0.25cm]{figures/logo/diamond.pdf}, while a diamond pickaxe \includegraphics[width=0.25cm]{figures/logo/diamond_pickaxe.pdf} can mine virtually any block in the game. Similarly, cultivating crops facilitates breeding various animals, each providing unique resources beyond sustenance. Drops from enemies also serve specific purposes, with some offering greater utility than others. By integrating resources from mining, farming, and breeding, players can enchant their equipment, further enhancing their capabilities. Additionally, collecting and crafting materials support construction, enabling players to create diverse structures. Beyond practical functions, such as building secure bases or farms, constructing personalized structures forms a significant aspect of the Minecraft experience. Figure~\ref{fig:example_long_horizon} illustrates an example of progression: crafting an iron sword \includegraphics[width=0.25cm]{figures/logo/iron_sword.pdf}.

\input{Supplementary/MGOA_data}

\input{Supplementary/picture/gen-data-pipeline}

\subsection{Observation and Action Spaces}

\textbf{Observation.}
In this paper, observation space of agent is completely consistent with human players. The agent only receives an RGB image with dimensions of $640 \times 360$ during the gameplay process, including the hotbar, health indicators, food saturation, and animations of the player's hands. It is worth helping the agent see more clearly in extremely dark environments, we have added a night vision effect for the agent, which increases the brightness of the environment during the night.

\noindent\textbf{Action Spaces.}
In MineRL \cite{guss2019minerl} environment, agent's action space is almost similar to human players. It consists of two parts: the mouse and the keyboard. The keypresses are responsible for controlling the movement of agents, such as jumping, forward, back, etc. The mouse movements are responsible for controlling the perspective of agents and the cursor movements when the GUI is opened. The left and right buttons of the mouse are responsible for attacking and using or placing items. In Minecraft, precise mouse movements are important when completing complex tasks that need open inventory or crafting table. In order to achieve both the same action space with MineDojo \cite{fan2022minedojo}, we abstract the craft and the smelt action into action space. The detailed action space is described in Table \ref{tab:action_space}.

\section{MGOA Dataset}
\label{sup:dataset}
In Minecraft, there is still a lack of sufficient high-quality goal-observation-action pairs to support the training of Optimus-2. To address this, we propose an automated dataset construction process aimed at creating high-quality Minecraft Goal-Observation-Action (MGOA) datasets. Through this method. MGOA contains 25,000 videos, providing about 30M goal-observation-action pairs. It contains 8 \textit{Atomic Tasks} across 5 tech levels: ‘Log \includegraphics[width=0.25cm]{figures/logo/wood.pdf}’, ‘Seed \includegraphics[width=0.25cm]{figures/logo/seeds.pdf}’, ‘Dirt \includegraphics[width=0.25cm]{figures/logo/dirt.pdf}’, ‘Stone \includegraphics[width=0.25cm]{figures/logo/cobblestone.pdf}’, ‘Iron \includegraphics[width=0.25cm]{figures/logo/iron_ore.pdf}’, ‘Gold \includegraphics[width=0.25cm]{figures/logo/gold_ingot.pdf}’, ‘Diamond \includegraphics[width=0.25cm]{figures/logo/diamond.pdf}’, ‘Redstone \includegraphics[width=0.25cm]{figures/logo/readstone.pdf}’. Note that the \textit{Atomic Tasks} in MGOA require minimal steps and can typically be completed within 2 $\sim$ 3 minutes. For instance, the task ‘Iron \includegraphics[width=0.25cm]{figures/logo/iron_ore.pdf}’ involves mining iron with a stone pickaxe, without the need to gather raw materials to craft the stone pickaxe. The statistics for the MGOA dataset is shown in Figure \ref{fig:mgoa_data}. We provide several examples of the dataset in the \texttt{MGOA\_samples} folder within the supplementary materials. We will release this dataset to contribute to the development of open-world agents within the community.

\subsection{Dataset Construction} 
\noindent\textbf{Pipeline.} Inspired by Li et al. \cite{li2024optimus}, we employed a prior policy (STEVE-1 \cite{lifshitz2024steve} in our work) to perform specific tasks in Minecraft, and recorded the corresponding videos and actions to generate goal-observation-action pairs. As illustrated in Figure~\ref{fig:dataset_pipeline}, we employed a custom script to extract item names from the Minecraft Wiki\footnote{https://minecraft.wiki/}. Using these item names, we queried GPT-4\footnote{https://openai.com/index/gpt-4-research/} with a predefined prompt template to generate task instructions, thereby constructing an Instruction Pool. The task instructions from the Instruction Pool serve as input to STEVE-1 \cite{lifshitz2024steve}, enabling it to interact with the environment to complete the tasks. During task execution, each frame and corresponding action were recorded and stored. To expedite data generation, we instantiated multiple policies and used parallelization to quickly produce large amounts of data. 

\noindent\textbf{Data Filtering.} We judged task success based on environmental feedback. For example, feedback like ``obtained new item, diamond axe'' indicated that the task ``craft a diamond axe'' was successfully completed. During the dataset generation process, we observed a significant amount of low-quality video data due to limitations in the policy's ability to follow instructions. Examples of low-quality data included task failures or task completion timeouts. To address this issue, we established two filtering criteria to ensure data quality: (1) only retaining data from successfully completed tasks, and (2) removing data for tasks that lasted longer than 2 minutes. These criteria allowed us to automatically filter out low-quality data, significantly reducing the cost of constructing the dataset. As a result, we obtained a high-quality MGOA dataset consisting of 25,000 samples.

\subsection{Comparison with Existing Datasets} 
Previous gameplay videos were primarily obtained through two methods below.

\noindent\textbf{Video Platform}: For example, MineDojo \cite{fan2022minedojo} collected game videos uploaded by human players on platforms such as YouTube and Twitter, combining the video content with corresponding titles or subtitles to form video-text pairs. However, this dataset lacked recorded actions. To address this, VPT \cite{vpt} used an Inverse Dynamics Model (IDM) to generate action sequences from the videos. However, the actions predicted by the IDM model are only approximations, which introduces a potential risk of misalignment between the frames and the corresponding actions.

\noindent\textbf{Human Contractors}: VPT \cite{vpt} hired human players to freely explore Minecraft and used the frames and actions to construct a video-action dataset. However, this dataset lacked explicit natural language instructions. To create goal-observation-action pairs, STEVE-1 \cite{lifshitz2024steve} used GPT-3.5 to generate specific task descriptions based on the gameplay, thereby integrating natural language instructions into the dataset. However, they provide only approximately 32k aligned goal-observation-action pairs, which remains a relatively scarce amount of data.

In addition, some work \cite{qin2023mp5,wang2024omnijarvis} have utilized GPT-4V to generate image captions, task planning, and reflections, thereby creating image-text pairs that form instruction-following datasets.

Distinct from the aforementioned datasets, the MGOA dataset directly captures agents performing specific tasks, offering clear natural language instructions with a one-to-one correspondence between observations and actions. Furthermore, through rigorous data filtering, redundant action sequences that do not contribute to task completion are excluded from MGOA. In addition, compared to the small-scale goal-observation-action datasets currently available, MGOA offers 25,000 videos, encompassing approximately 30 million goal-observation-action pairs. This dataset is not only significantly larger but also highly scalable in an automated manner.

\input{Supplementary/atomic-task}
\input{Supplementary/long-horizon}
\section{Training Details}
\label{sup:train}
\subsection{Training Pipeline}
One of the key factors in implementing our proposed method lies in the efficient alignment of language with the observation-action sequence, and subsequently translating language space into the action space. To tackle this problem, we adopt a two-phase training approach. First, we align language with the observation-action sequence via behavior pre-training. Then, we transform the language space into the action space through action fine-tuning. 

\input{Supplementary/table/tb_hyperparameter}
\noindent\textbf{Behavior Pre-training}: During the pre-training phase, we integrated the Vision-guided Behavior Encoder into the model. We used OpenAI Contractor Dataset \cite{vpt} and a subset of MGOA as training data, which comprised approximately 5,000 videos. To balance efficiency and effectiveness, we freeze the visual encoder, then tune the Vision-guided Behavior Encoder along with a large language model (LoRA \cite{hu2021lora}). During pre-training, we set the learning rate to 0.0001 and trained for 5 epochs. The hyperparameter settings are shown in Table \ref{tb:hyper}.

\noindent\textbf{Action Fine-tuning}: During the fine-tuning phase, we adapted the general MLLM DeepSeek-VL-1.3B \cite{lu2024deepseek} to the Minecraft environment, transitioning the model's output space from language to low-level actions. We fine-tuned it using OpenAI Contractor Dataset \cite{vpt} and MGOA, which comprises approximately 20,000 videos. In this phase, we freeze the Vision-guided Behavior Encoder, visual encoder, and large language model (LoRA), and only fine-tuned the action head. During fine-tuning, we set the learning rate to 0.00004 and train for 10 epochs. The hyperparameter settings are shown in Table \ref{tb:hyper}.

\subsection{Implementation Details}
For the planner, we follow Li et al. \cite{li2024optimus}, employing Multimodal Hybrid Memory empowered GPT-4V for planning and reflection. For the policy, we train the GOAP through the above pipeline. All experiments were conducted on 8x NVIDIA L40 GPUs. For the MGOA dataset, data collection and filtering were conducted in parallel, taking approximately 7 days. Training required around 2 days, while inference and evaluation on atomic tasks, long-horizon tasks, and open-ended instruction tasks took approximately 4 days.

\section{Benchmark}
\label{sup:evaluate}
\subsection{Evaluation Tasks}
The evaluation tasks are divided into three categories: \textit{Atomic Tasks}, \textit{Long-horizon Tasks}, and \textit{Open-ended Instruction Tasks}. For each task, the agent's environment is randomly initialized each time, and every task is executed at least 30 times. For \textit{Atomic Tasks}, we follow the setting of prior work \cite{lifshitz2024steve,wang2024omnijarvis}, which requires the agent to execute the task within 2 minutes. We then report the average reward for the task, defined as the number of items obtained.
 For \textit{Open-ended Instruction Tasks} and \textit{Long-horizon Tasks}, we report the average success rate (SR) for each task. 

\textit{Atomic Tasks.} As shown in Figure \ref{fig:example_atomic}, Atomic Tasks are short-term skills in Minecraft, such as ``chop a tree to get logs \includegraphics[width=0.25cm]{figures/logo/wood.pdf}'', ``mine dirt \includegraphics[width=0.25cm]{figures/logo/dirt.pdf}'', ``collect seeds \includegraphics[width=0.25cm]{figures/logo/seeds.pdf}'', and ``dig down to mine stone \includegraphics[width=0.25cm]{figures/logo/cobblestone.pdf}'', etc.

\textit{Long-horizon Tasks.} As shown in Figure \ref{fig:example_long_horizon}, Long-Horizon Tasks are a sequence of \textit{Atomic Tasks}. For example, ``craft an iron sword from scratch'' requires completing the atomic tasks of ``chop 7 logs'', ``craft 21 planks'', ``craft 5 sticks'', ``craft 1 crafting table'', and so on. These \textit{Atomic Tasks} are interdependent, meaning that the failure of any single atomic task will result in the failure of the entire \textit{Long-horizon Task}.

\textit{Open-ended Instruction Tasks.} Open-Ended Instruction Tasks are not limited to predefined text formats; rather, they involve flexible language directives that prompt the agent to accomplish long-horizon tasks. These tasks evaluate the agent’s capacity to interpret and execute instructions expressed in open-ended natural language. We selected Torch \includegraphics[width=0.25cm]{figures/logo/torch.pdf}, Rail \includegraphics[width=0.3cm]{figures/logo/rail.pdf}, Golden Shovel \includegraphics[width=0.3cm]{figures/logo/Golden_Shovel.pdf}, Diamond Pickaxe \includegraphics[width=0.3cm]{figures/logo/diamond_pickaxe.pdf}, and Compass \includegraphics[width=0.3cm]{figures/logo/compass.pdf} as evaluation tasks. Instruction for each task are shown in Table \ref{tab:ins_craft_torch}, Table \ref{tab:ins_craft_rail}, Table \ref{tab:ins_craft_golden_shovel}, Table \ref{tab:ins_craft_diamond_pickaxe} and Table \ref{tab:ins_craft_compass}.

\subsection{Baselines}
In this section, we provide a brief overview of existing Minecraft agents and compare them with our proposed Optimus-2. Current agents can be broadly categorized into two types: policy-based agents and planner-policy agents.

\noindent\textbf{Policy-based Agents}. Policy-based agents \cite{vpt,cai2023groot,lifshitz2024steve,fan2022minedojo,cai2023open} refer to those trained through reinforcement learning or imitation learning, capable of completing atomic tasks within Minecraft. However, due to limitations in instruction understanding and reasoning abilities, they struggle to accomplish long-horizon tasks.

\noindent\textbf{Planner-Policy Agents}. Planner-policy agents \cite{wang2023describe,qin2023mp5,li2024auto,wang2023jarvis,li2024optimus,wang2024omnijarvis} refer to non-end-to-end architectures that utilize a MLLM (Multi-Layered Language Model) as a planner to decompose complex instructions into a sequence of sub-goals executable by a policy. While significant progress has been made, the current performance bottleneck stems from the policy's ability to effectively understand and execute the sub-goals generated by the planner.

\noindent\textbf{Comparison with Existing Agents}. As a core contribution of this work, we propose a novel Goal-Observation-Action Conditioned Policy, GOAP. It integrates two key components: an Action-Guided Behavior Encoder for modeling observation-action sequences, and an MLLM for aligning sub-goals with these sequences. Leveraging the MLLM's advanced understanding of open-ended instructions, GOAP demonstrates superior instruction-following capabilities compared to existing policies. On top of GOAP, the proposed agent, Optimus-2, exhibits superior performance in long-horizon tasks, outperforming the current state-of-the-art across all seven task groups.

\section{Experimental Results}
\label{sup:exp}
In this section, we report the experimental results of Optimus-2 on each \textit{Long-horizon task}.

\subsection{Results on Long-horizon Task}
In this section, we report the results of Optimus-2 on each \textit{Long-horizon Task}, with details including task name, numbers of sub-goals, success rate (SR), and eval times. As shown in Tables \ref{tab:wooden_result} and \ref{tab:gold_result}, Optimus-2 demonstrates superior performance across all 67 \textit{Long-horizon Tasks}. Since Optimus-2 is randomly initialized in arbitrary environments for each task execution, the experimental results also highlight its generalization capability across diverse environments.

\section{Case Study}
\label{sup:case}
In this section, we provide additional cases to illustrate the differences in the ability of VPT (text) \cite{vpt}, STEVE-1 \cite{lifshitz2024steve}, and Optimus-2 to perform \textit{Open-ended Instruction Tasks}. We provide different open-ended instructions requiring the agent to perform tasks across various biomes. As shown in Figure \ref{fig:case1}, Figure \ref{fig:case2}, and Figure \ref{fig:case3}, Optimus-2 effectively completes all tasks, while VPT (text) and STEVE-1 fail due to limitations in language understanding and multimodal perception capabilities. Moreover, we provide several demo videos of Optimus-2 performing long-horizon tasks in the \texttt{Optimus2\_videos} folder within the supplementary materials.

\input{Supplementary/open_example}

\input{Supplementary/table/optimus1_benchmark_result/wooden_result}

\input{Supplementary/table/optimus1_benchmark_result/golden_result}

\input{Supplementary/case_study}

\clearpage

%% file: Supplementary/picture/behavior.tex
\begin{figure}[htbp]
    \centering
    \includegraphics[width=0.5\textwidth]{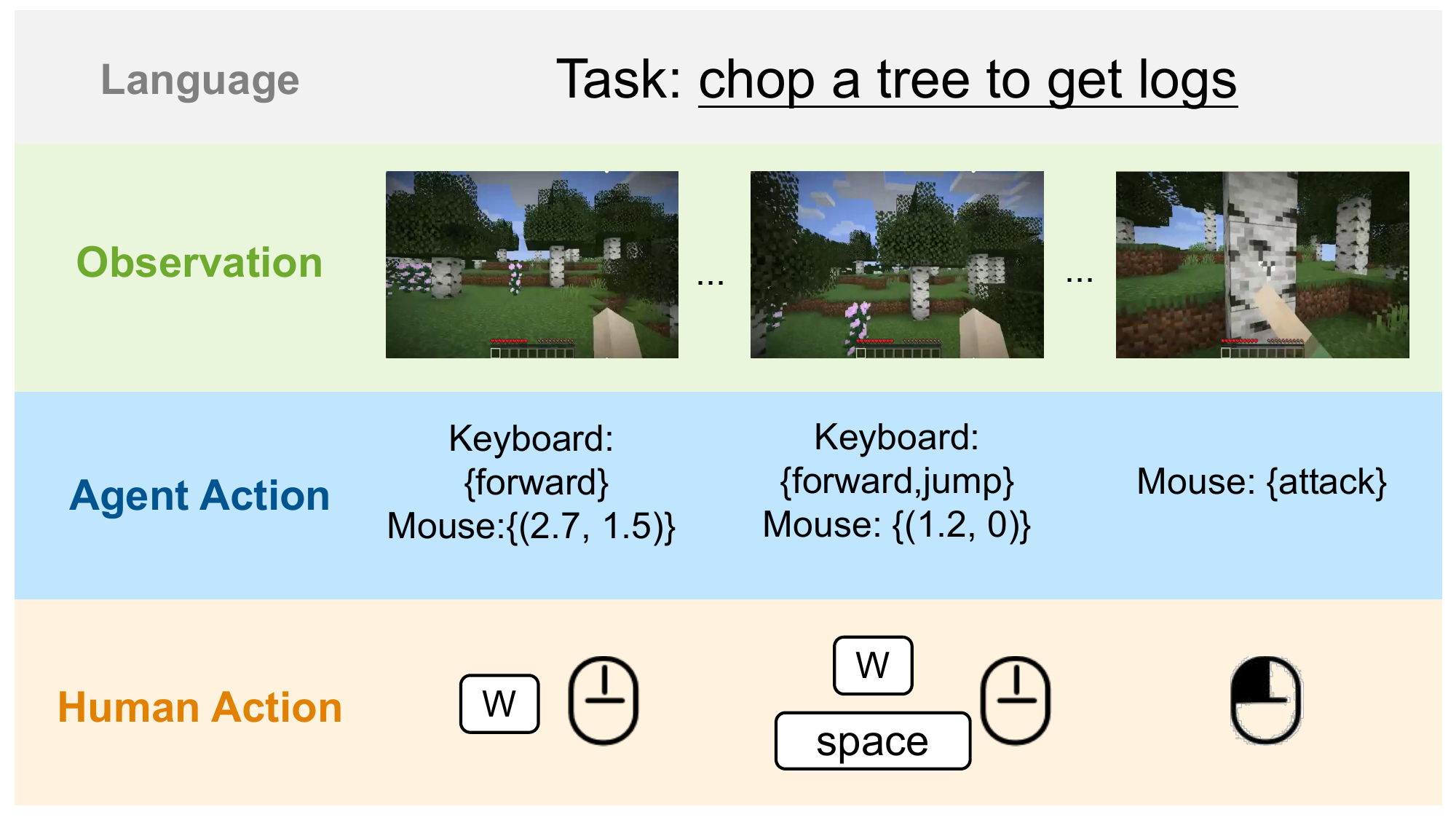}
    \caption{Illustration of behavior patterns of both human and agents in Minecraft.}
    \label{fig:behavior}
\end{figure}

%% file: Supplementary/table/action_space.tex
\begin{table*}[htbp]
\centering
\caption{Action space of agent in Minecraft.}
\label{tab:action_space}
\resizebox{\textwidth}{!}{%
\renewcommand\arraystretch{1.2}
\begin{tabular}{llllcc}
\toprule[1.5pt]
\textbf{Index} & \textbf{Agent Action}      & \textbf{Human Action} & \textbf{Description}                               \\ \hline
1              & Forward              & key W                 & Move forward.                                      \\
2              & Back                 & key S                 & Move back.                                         \\
3              & Left                 & key A                 & Strafe left.                                       \\
4              & Right                & key D                 & Strafe right.                                      \\
5              & Jump                 & key Space             & Jump. When swimming, keeps the player afloat.      \\
6              & Sneak                & key left Shift        & Slowly move in the current direction of movement.  \\
7              & Sprint               & key left Ctrl         & Move quickly in the direction of current movement. \\
8  & Attack & left Button  & Destroy blocks (hold down); Attack entity (click once).                     \\
9  & Use    & right Button & Place blocks, entity, open items or other interact actions defined by game. \\
10             & hotbar {[}1-9{]}     & keys 1-9              & Selects the appropriate hotbar item.               \\
11             & Open/Close Inventory & key E                 & Opens the Inventory. Close any open GUI.           \\
12 & Yaw    & move Mouse X & Turning; aiming; camera movement.Ranging from -180 to +180.                 \\
13 & Pitch  & move Mouse Y & Turning; aiming; camera movement.Ranging from -180 to +180.                 \\
14             & Craft                & -                     & Execute a crafting recipe to obtain new item       \\
15             & Smelt                & -                     & Execute a smelting recipe to obtain new item.      \\ 
\bottomrule[1.5pt]
\end{tabular}%
}
\end{table*}

%% file: Supplementary/MGOA_data.tex
\begin{figure}[htbp]
    \centering
    \includegraphics[width=0.5\textwidth]{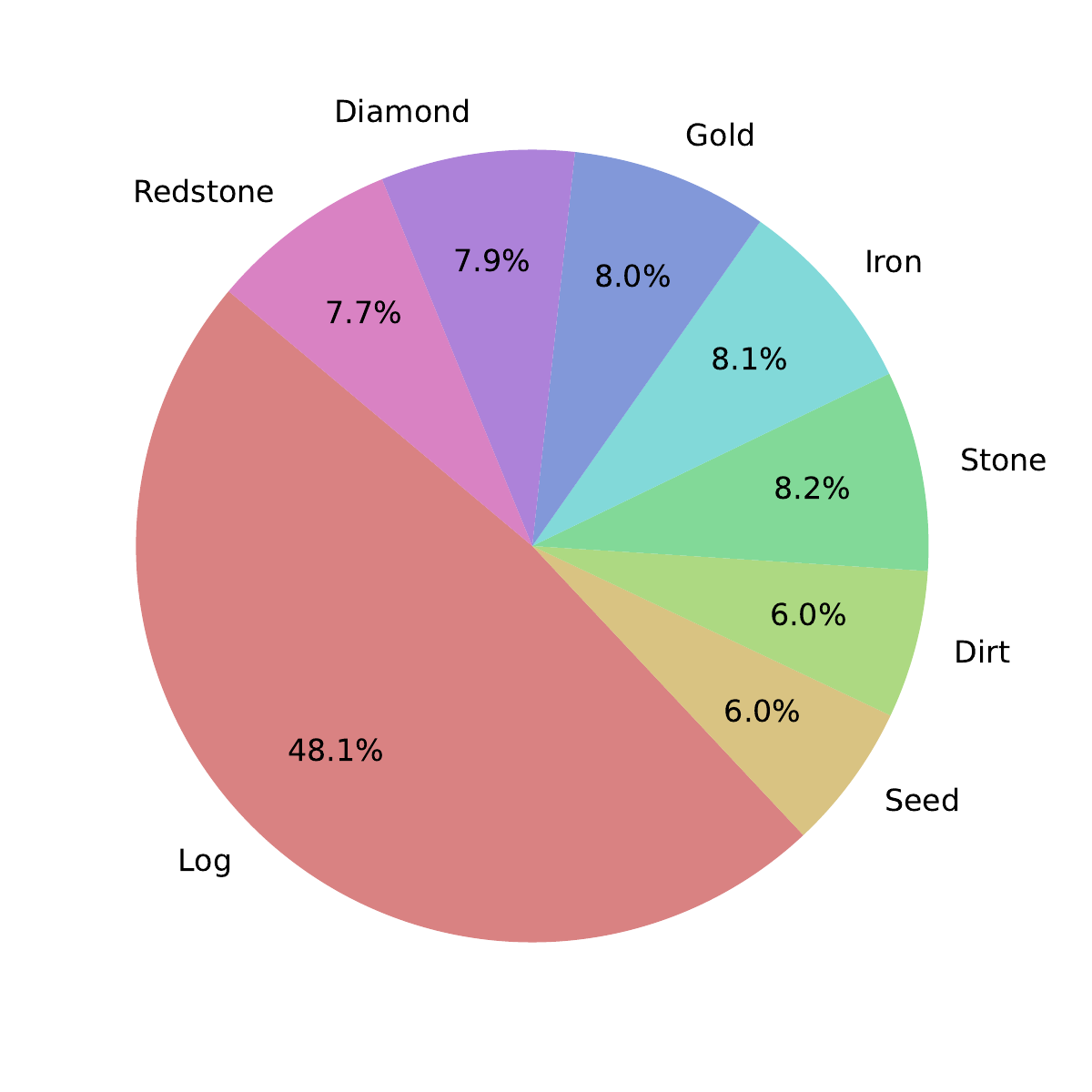}
    \vspace{-10pt}
    \caption{Statistical information on MGOA dataset. It contains 8 \textit{Atomic Tasks}: ‘Log \includegraphics[width=0.25cm]{figures/logo/wood.pdf}’, ‘Seed \includegraphics[width=0.25cm]{figures/logo/seeds.pdf}’, ‘Dirt \includegraphics[width=0.25cm]{figures/logo/dirt.pdf}’, ‘Stone \includegraphics[width=0.25cm]{figures/logo/cobblestone.pdf}’, ‘Iron \includegraphics[width=0.25cm]{figures/logo/iron_ore.pdf}’, ‘Gold \includegraphics[width=0.25cm]{figures/logo/gold_ingot.pdf}’, ‘Diamond \includegraphics[width=0.25cm]{figures/logo/diamond.pdf}’, ‘Redstone \includegraphics[width=0.25cm]{figures/logo/readstone.pdf}’.}
    \label{fig:mgoa_data}
\end{figure}

%% file: Supplementary/picture/gen-data-pipeline.tex
\begin{figure*}[htbp]
    \centering
    \includegraphics[width=0.9\textwidth]{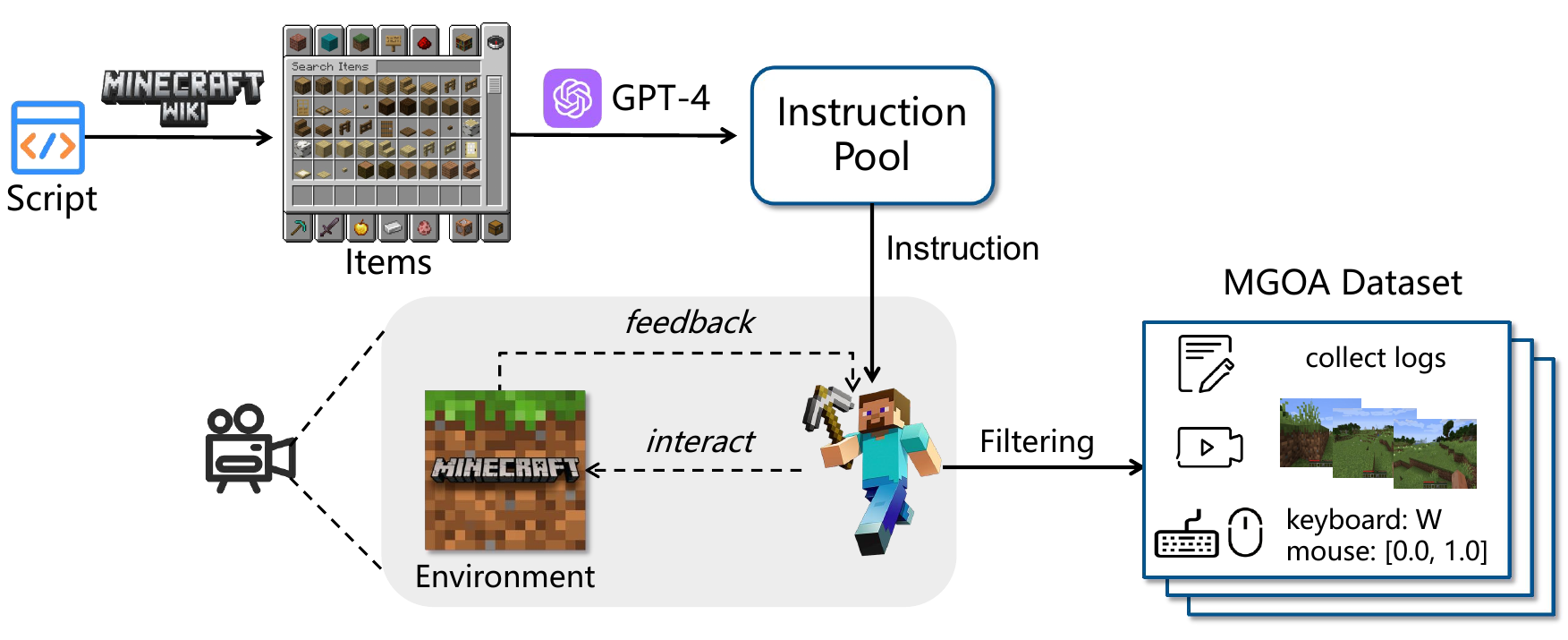}
    \caption{The pipeline for generating the MGOA dataset. First, we extracted item names from the Minecraft Wiki and employed GPT-4 to generate corresponding instructions. These instructions were then provided as input to STEVE-1, enabling it to interact with the environment to accomplish the tasks. During task execution, each observation was paired with its corresponding action, resulting in the creation of goal-observation-action pairs.}
    \label{fig:dataset_pipeline}
\end{figure*}

%% file: Supplementary/atomic-task.tex
\begin{figure*}[htbp]
    \subfloat[chop a tree to get logs \label{atomic_example:fig1}]{%
      \includegraphics[width=0.5\textwidth]{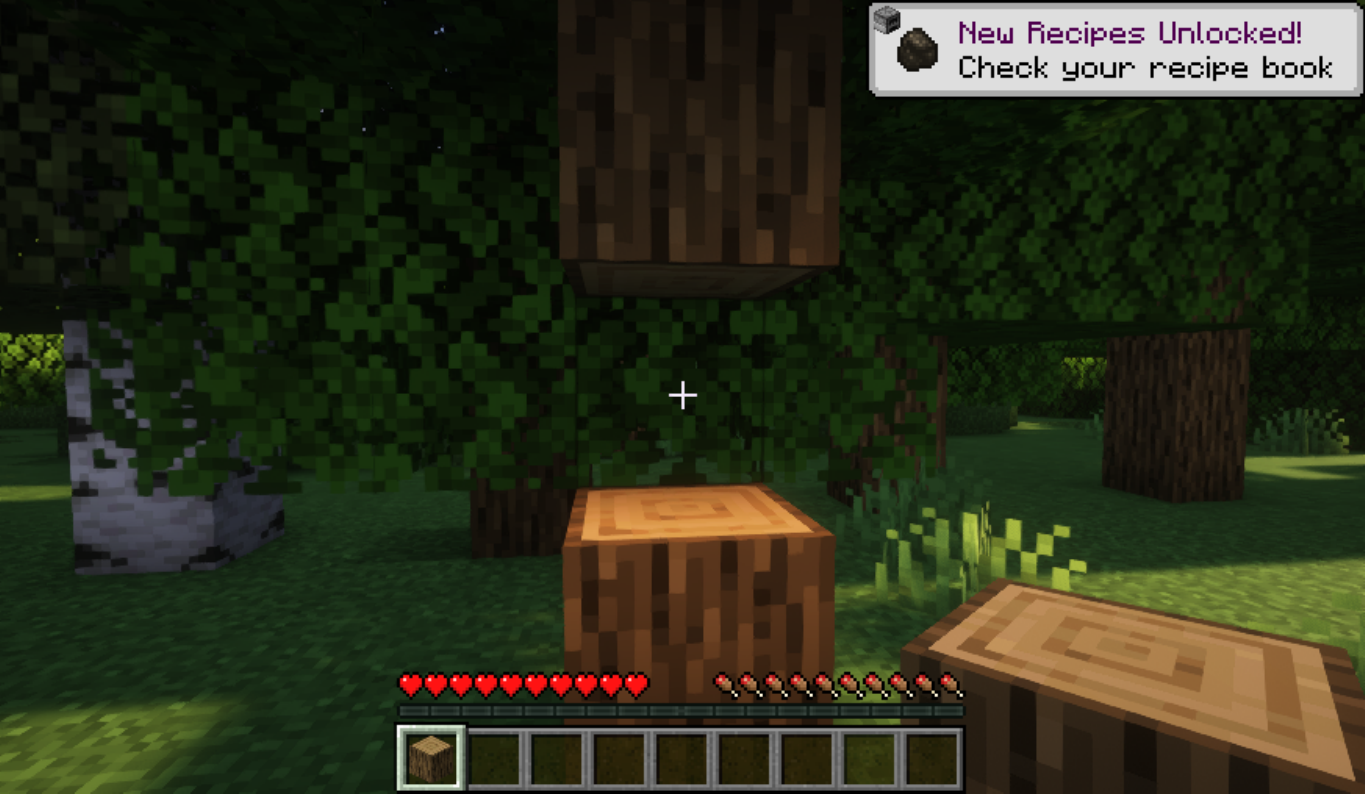}
    }
    \hfill
    \subfloat[mine dirt\label{atomic_example:fig2}]{%
      \includegraphics[width=0.5\textwidth]{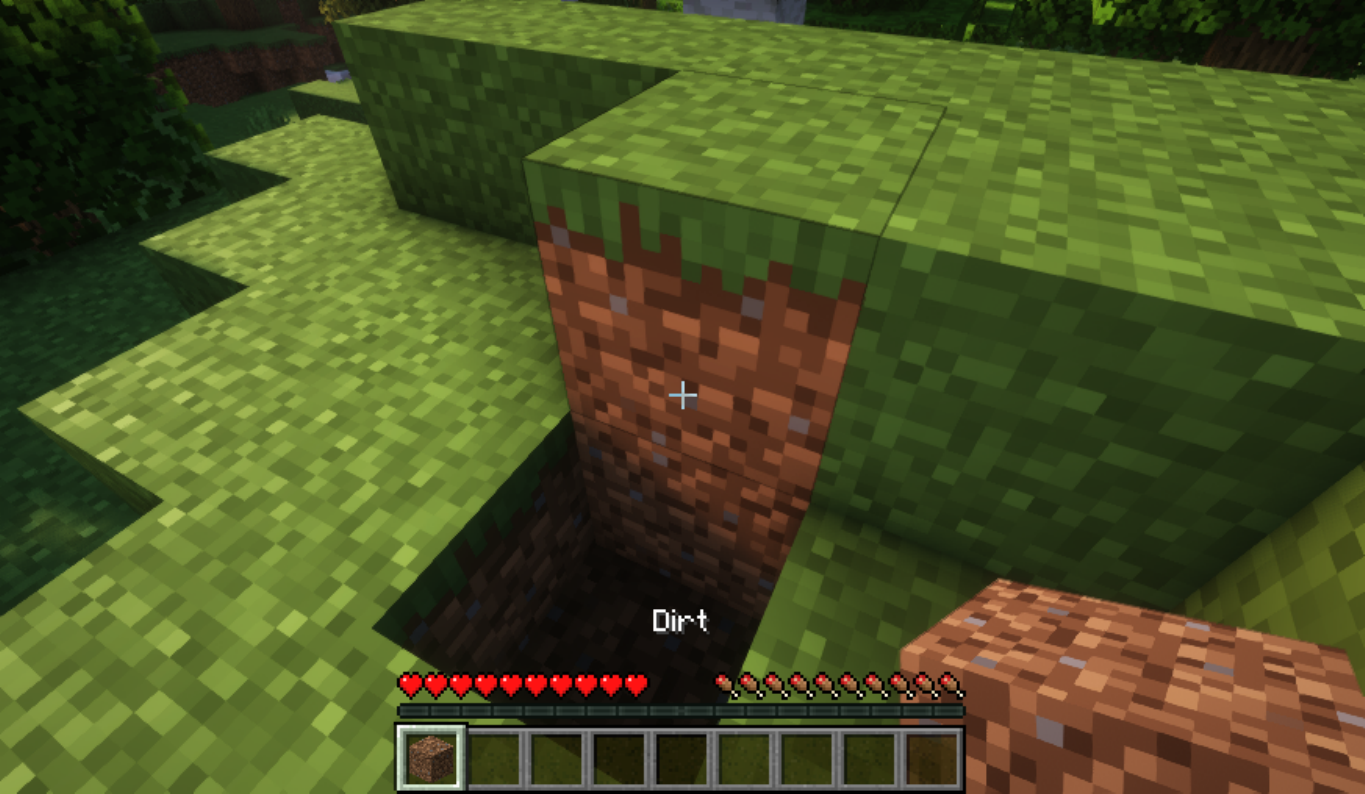}
    }
    \hfill
    \subfloat[collect seeds\label{atomic_example:fig3}]{%
      \includegraphics[width=0.5\textwidth]{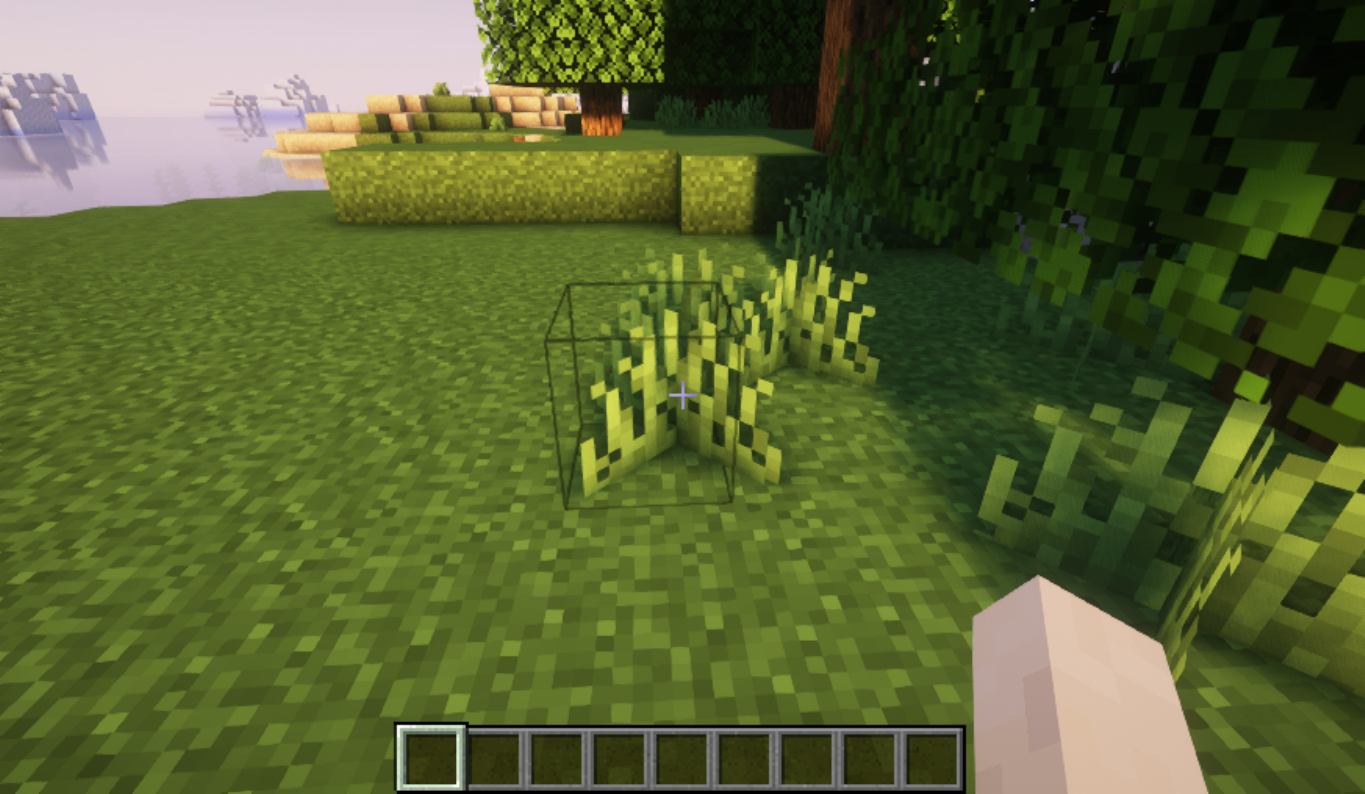}
    }
    \hfill
    \subfloat[dig down to mine stone\label{atomic_example:fig11}]{%
      \includegraphics[width=0.5\textwidth]{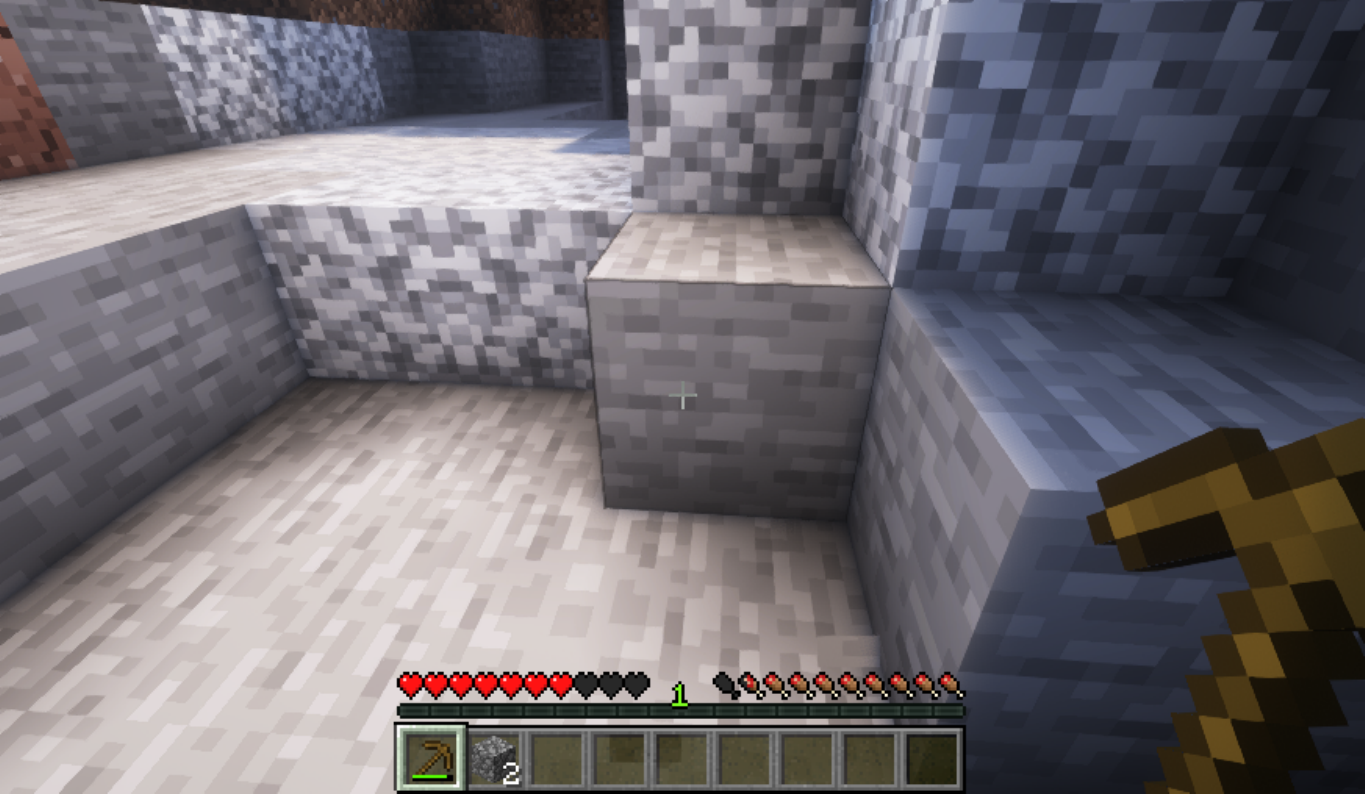}
    }
    \caption{Examples of \textit{Atomic Task}. The agent must follow the instructions to collect resources. These four tasks represent the basic capabilities of the agent. The more resources are collected, the stronger the basic capabilities of the agent will be.
 }
    \label{fig:example_atomic}
\end{figure*}

%% file: Supplementary/long-horizon.tex
\begin{figure*}[htbp]
    \subfloat[Chop $7$ logs\label{example:fig1}]{%
      \includegraphics[width=0.3\textwidth]{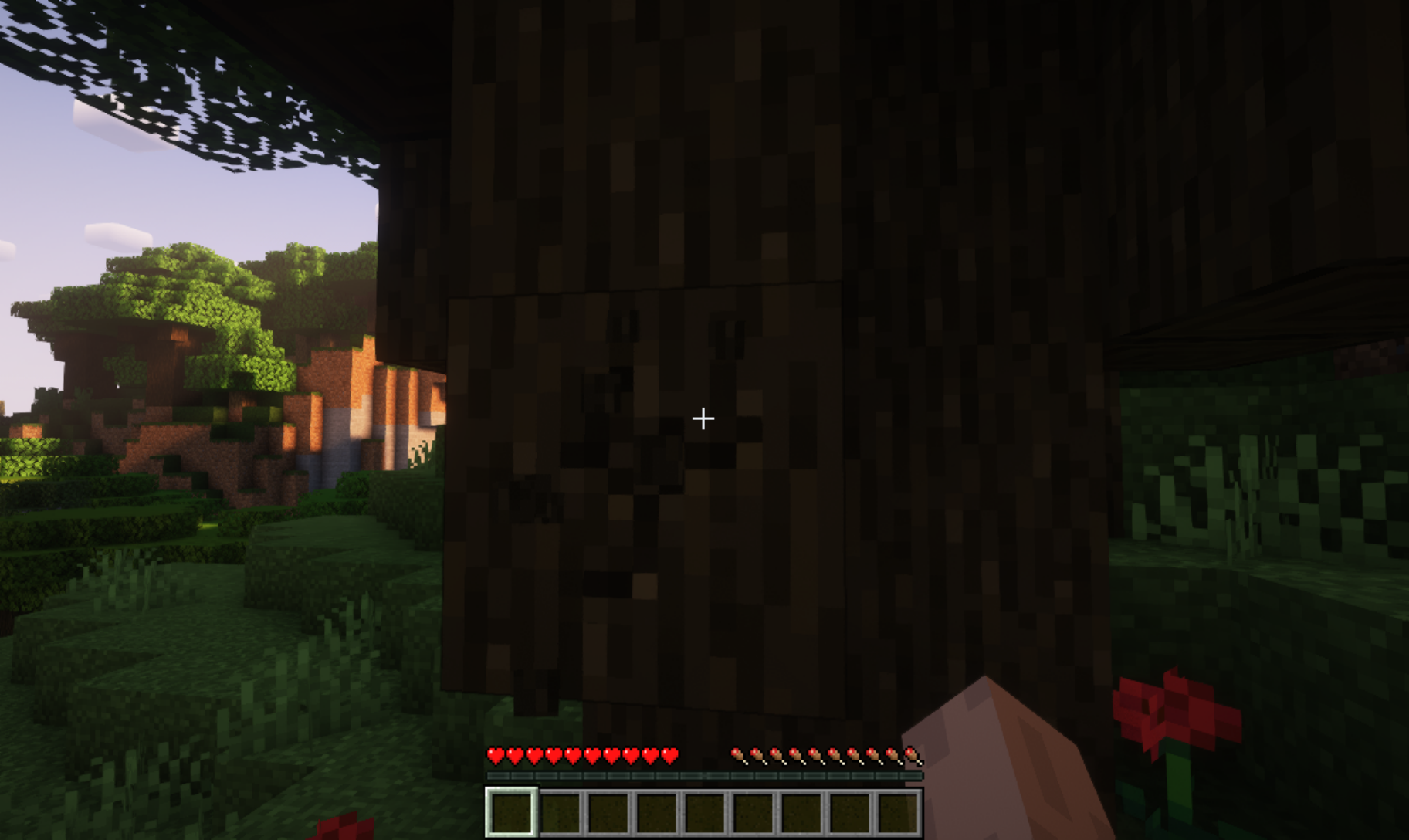}
    }
    \hfill
    \subfloat[Craft $21$ planks\label{example:fig2}]{%
      \includegraphics[width=0.3\textwidth]{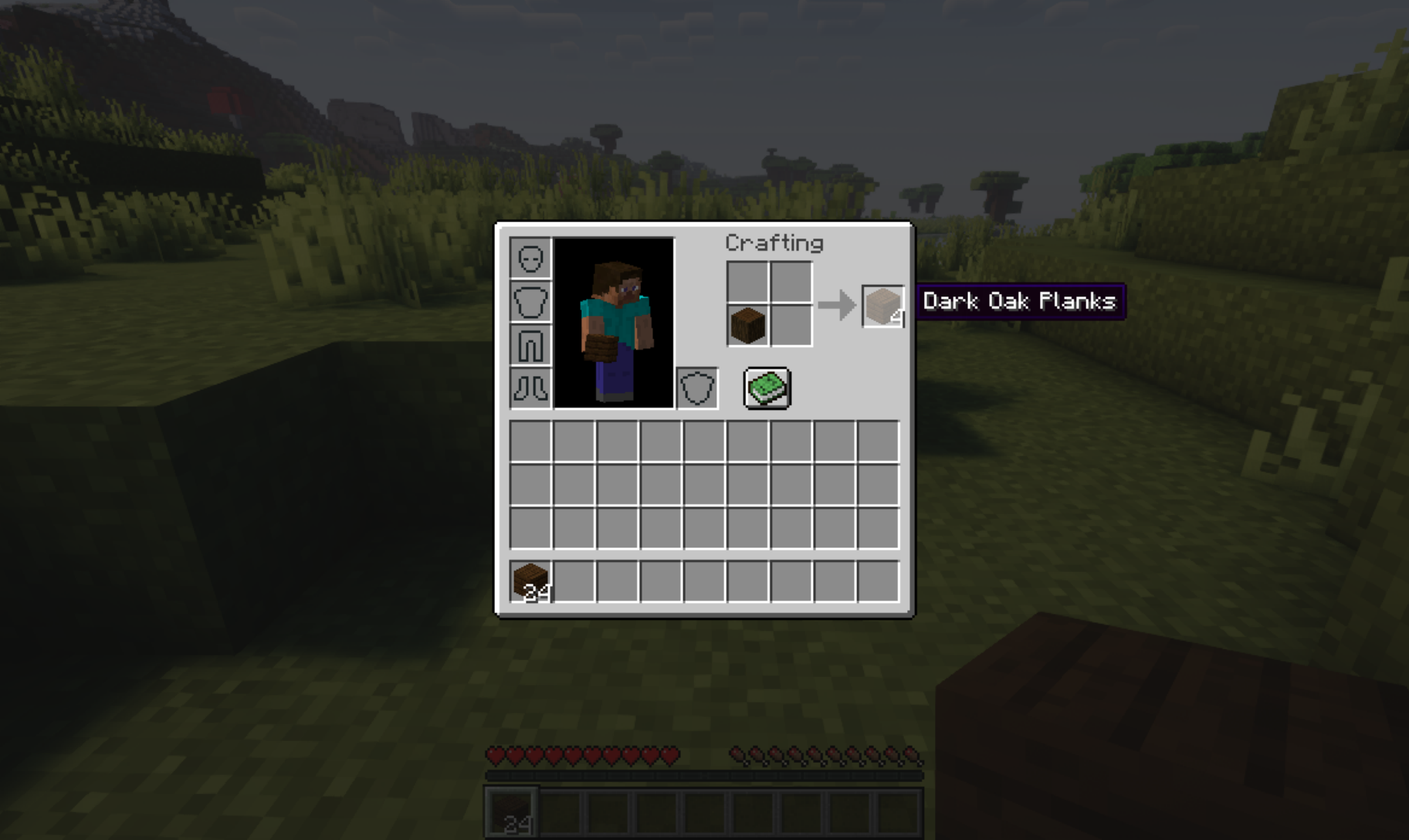}
    }
    \hfill
    \subfloat[Craft $5$ sticks\label{example:fig3}]{%
      \includegraphics[width=0.3\textwidth]{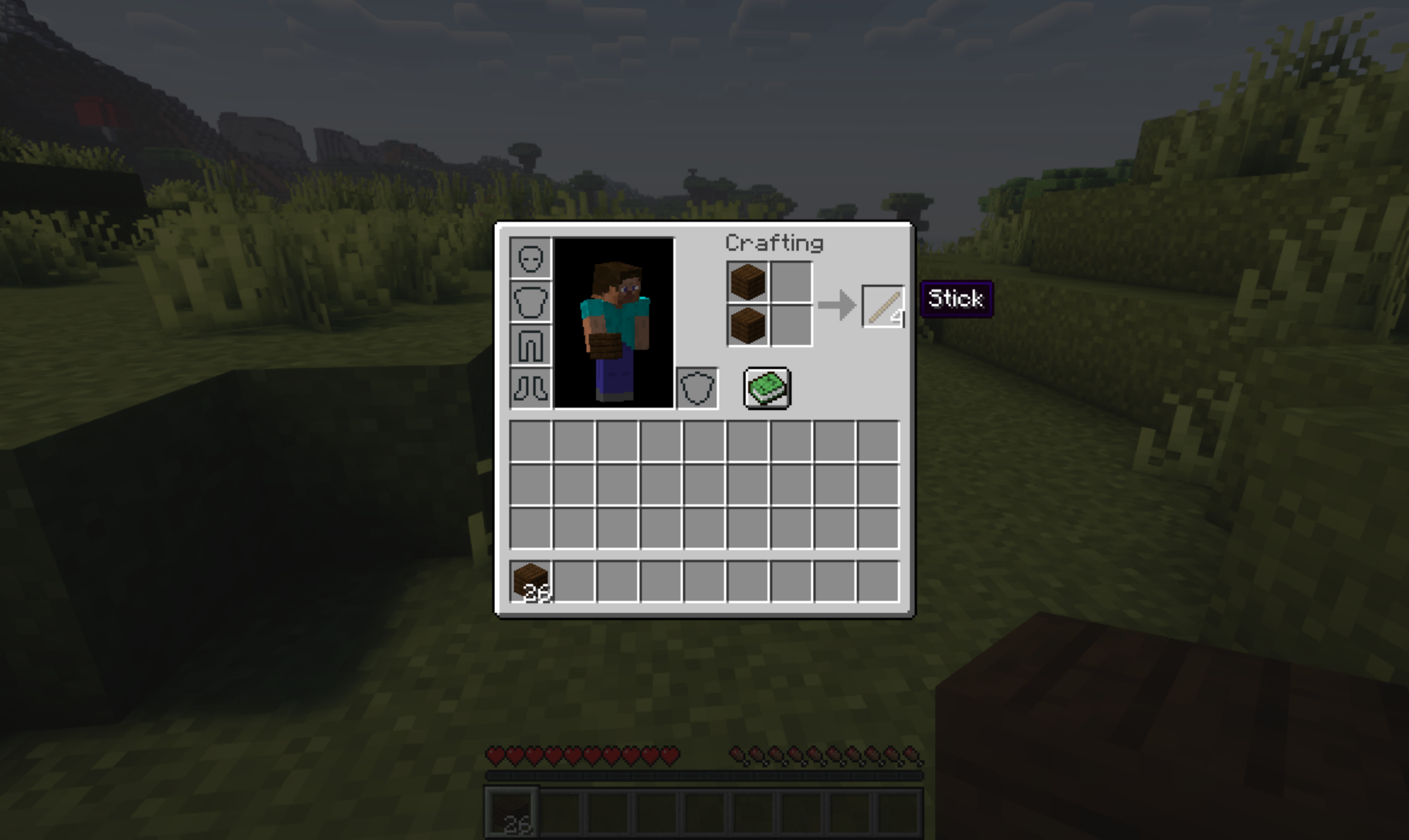}
    }
    \hfill
    \subfloat[Craft $1$ crafting table\label{example:fig11}]{%
      \includegraphics[width=0.3\textwidth]{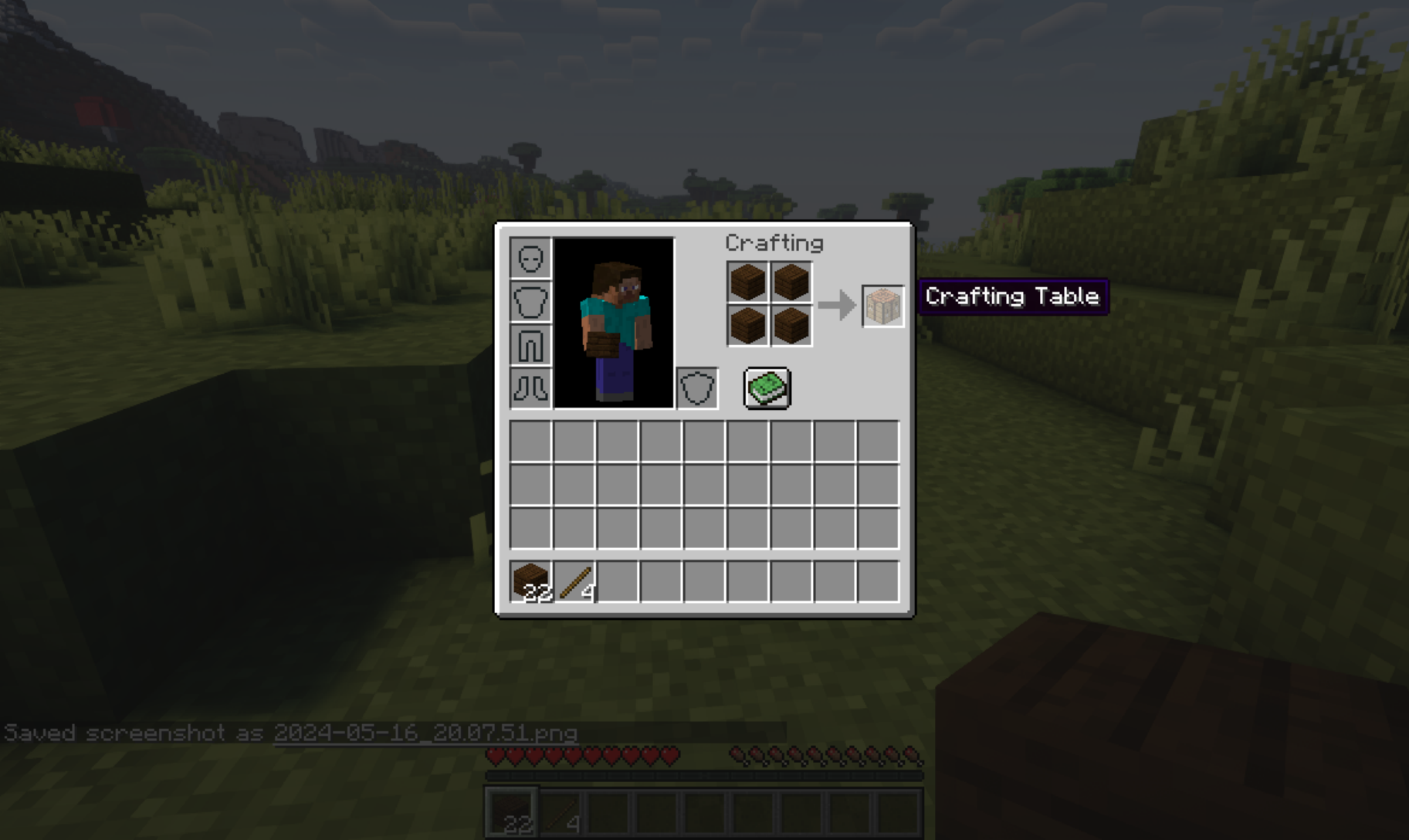}
    }
    \hfill
    \subfloat[Craft $1$ wooden pickaxe\label{example:fig4}]{%
      \includegraphics[width=0.3\textwidth]{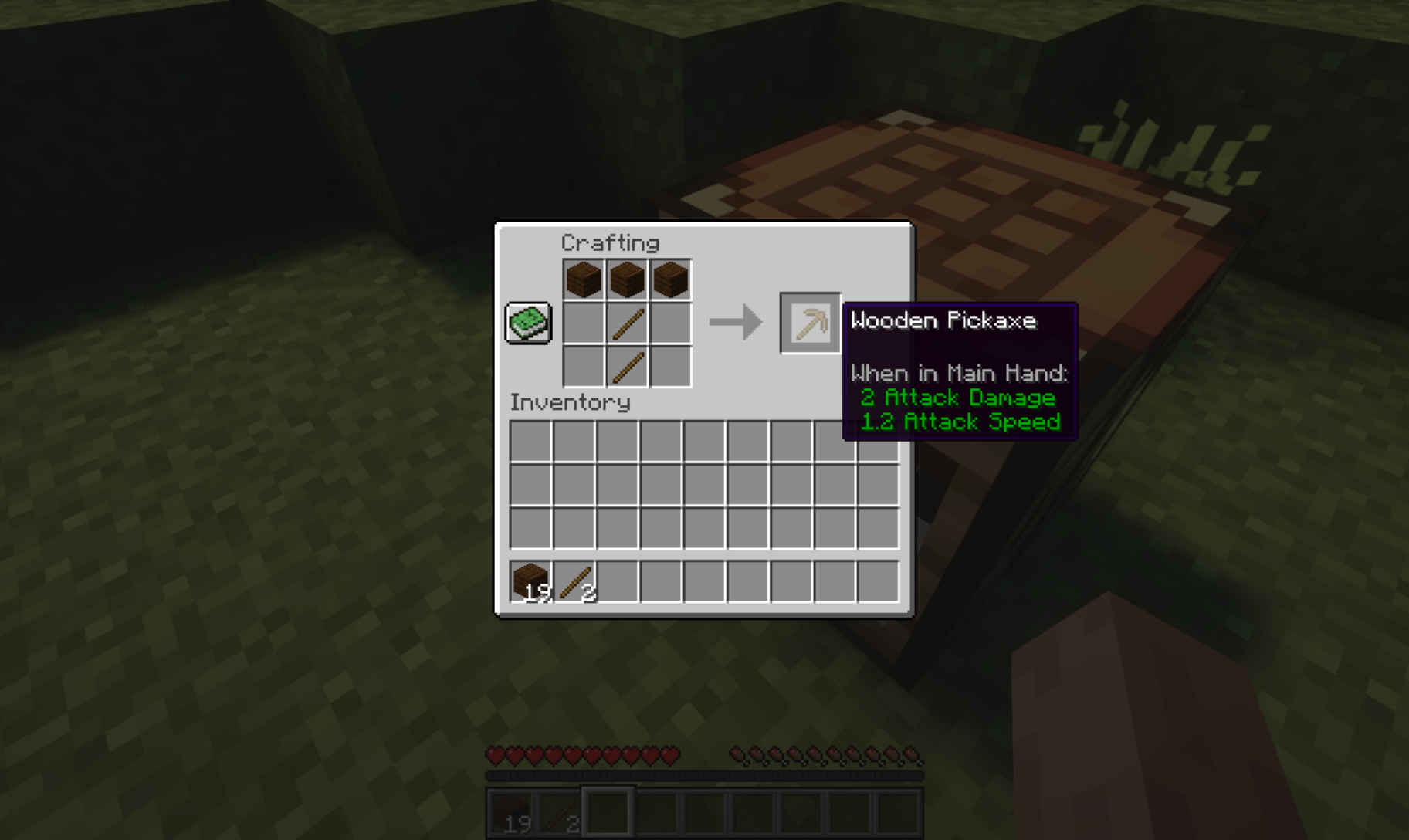}
    }
    \hfill
    \subfloat[Mine $11$ cobblestone\label{example:fig5}]{%
      \includegraphics[width=0.3\textwidth]{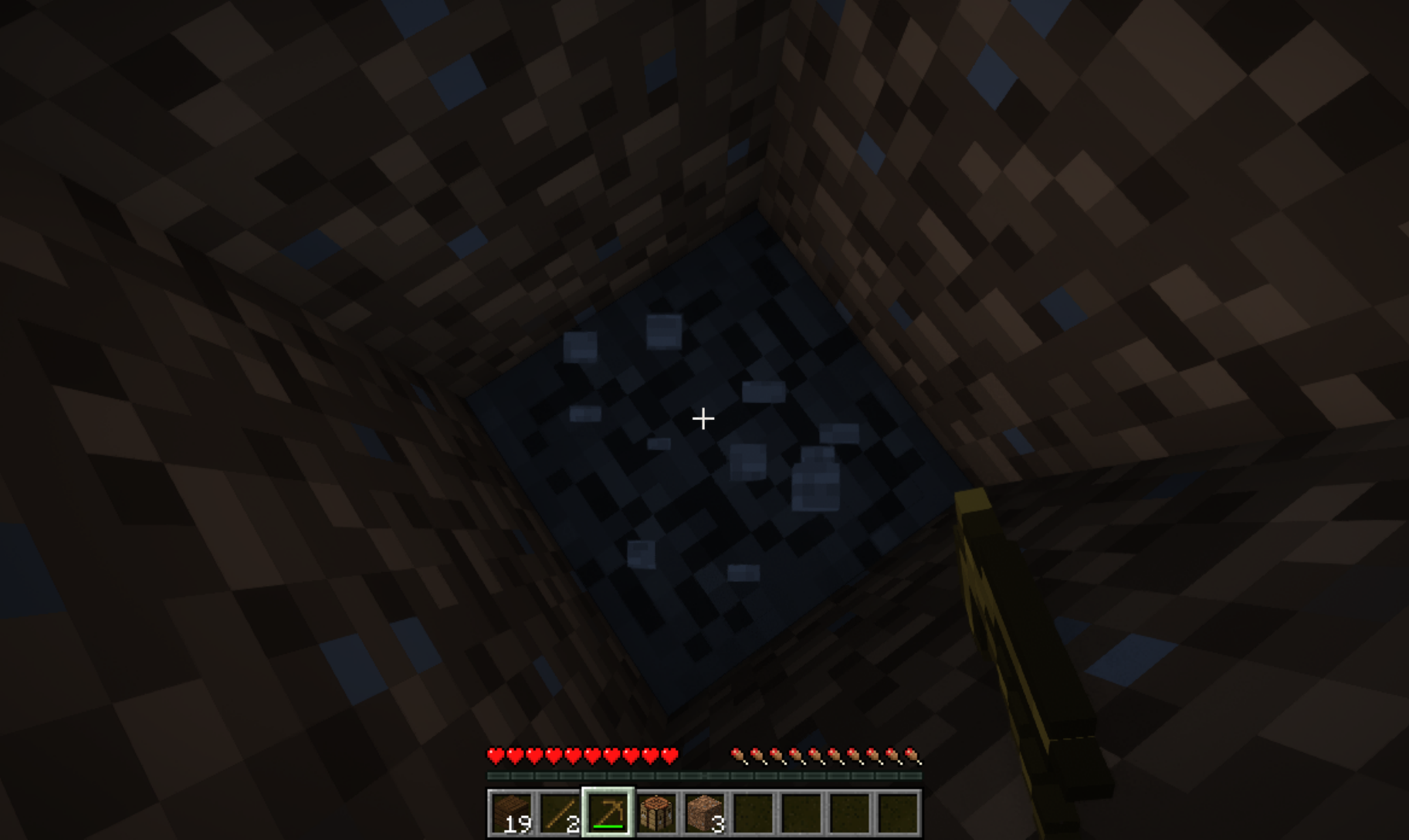}
    }
    \hfill
    \subfloat[Craft $1$ furnace\label{example:fig6}]{%
      \includegraphics[width=0.3\textwidth]{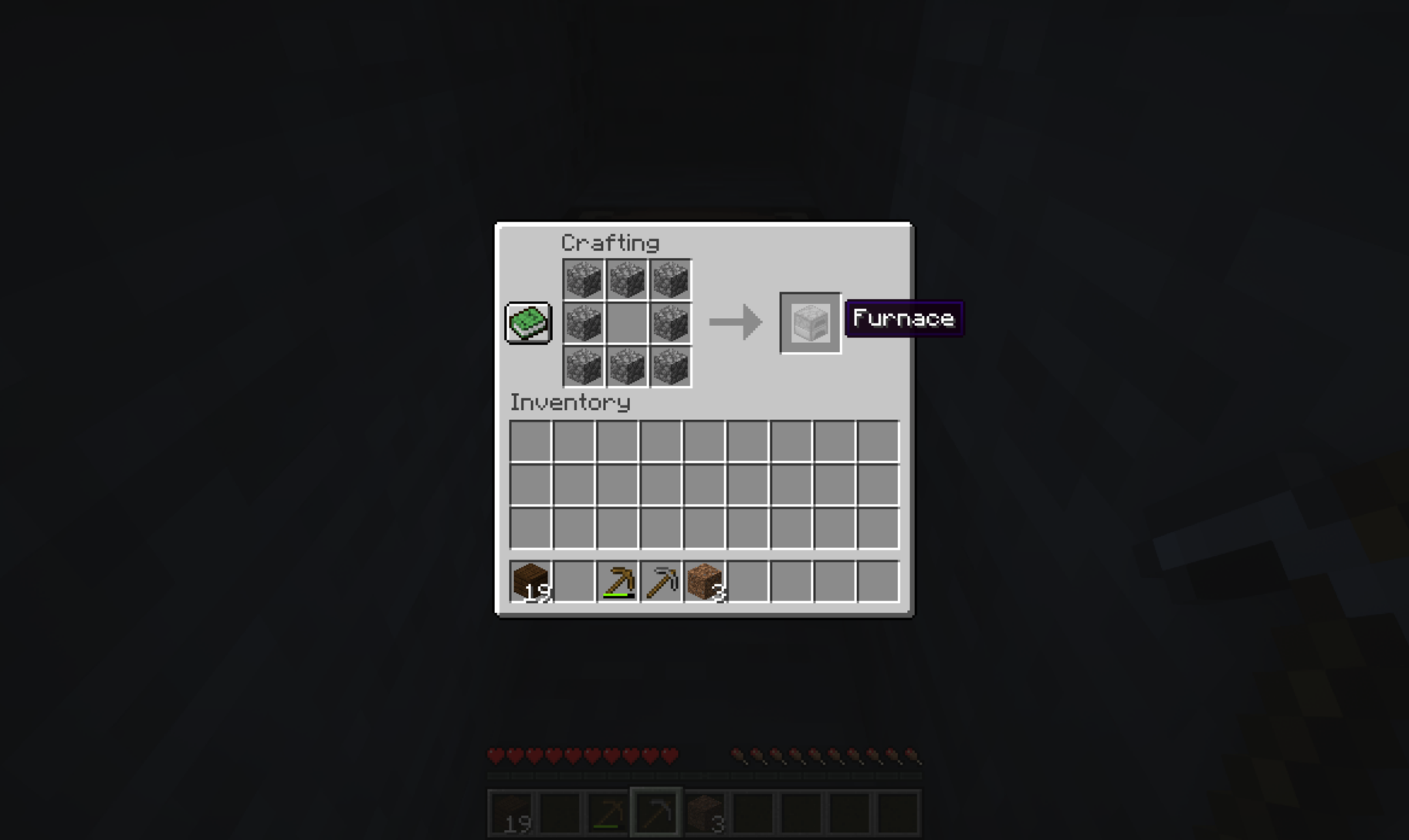}
    }
    \hfill
    \subfloat[Craft $1$ stone pickaxe\label{example:fig7}]{%
      \includegraphics[width=0.3\textwidth]{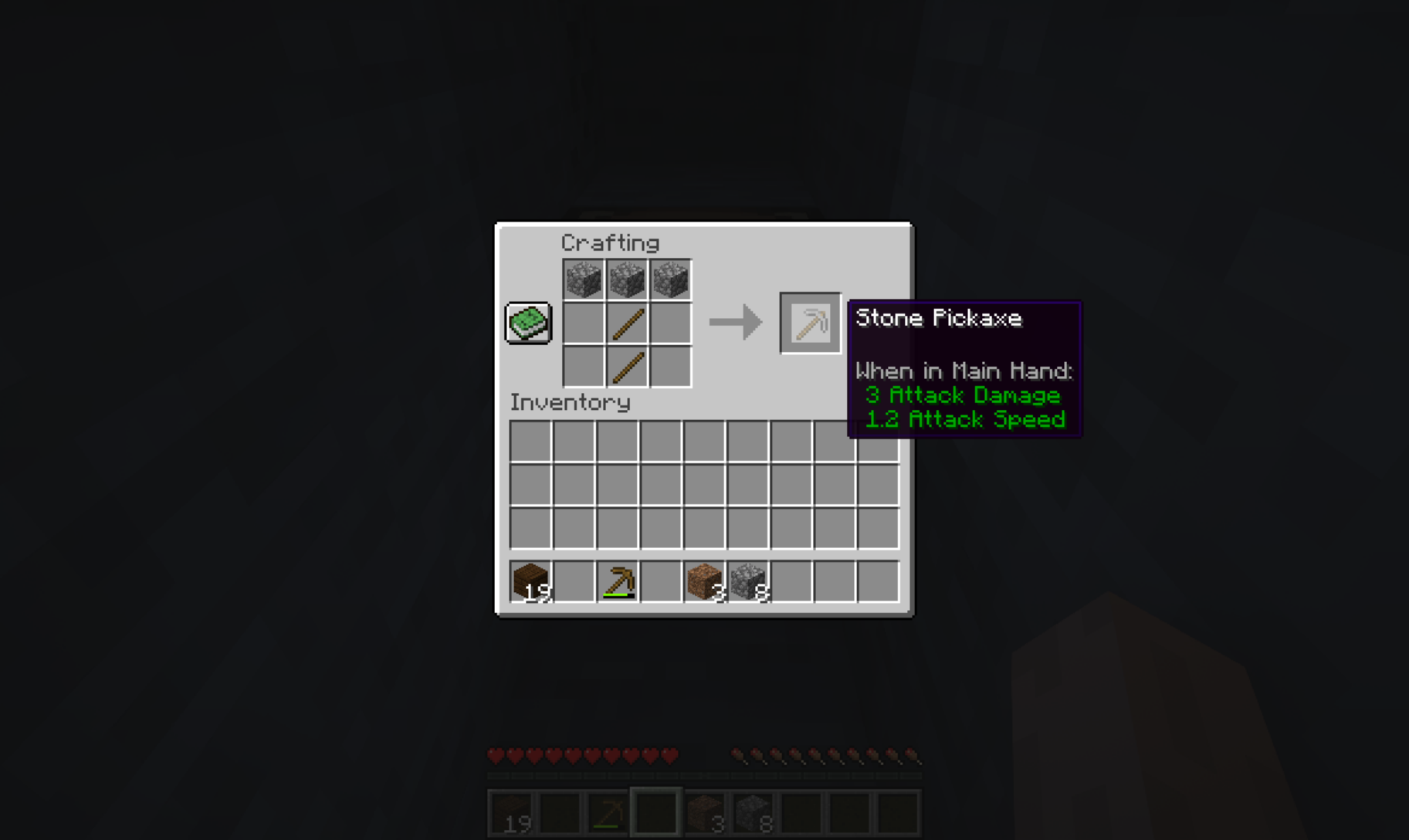}
    }
    \hfill
    \subfloat[Dig down more deeper to find iron ore\label{example:fig12}]{%
      \includegraphics[width=0.3\textwidth]{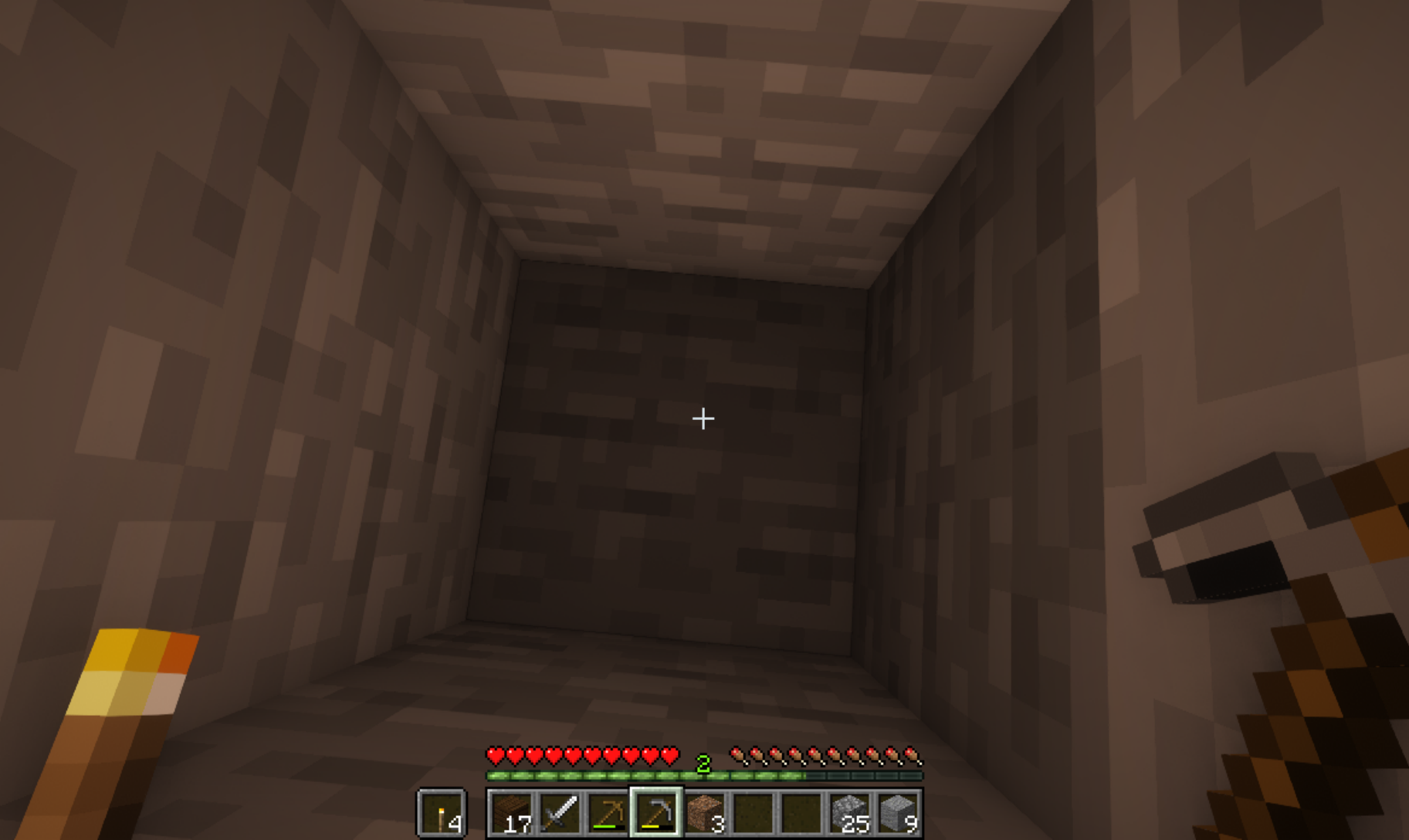}
    }
    \hfill
    \subfloat[Mine $2$ iron ores\label{example:fig8}]{%
      \includegraphics[width=0.3\textwidth]{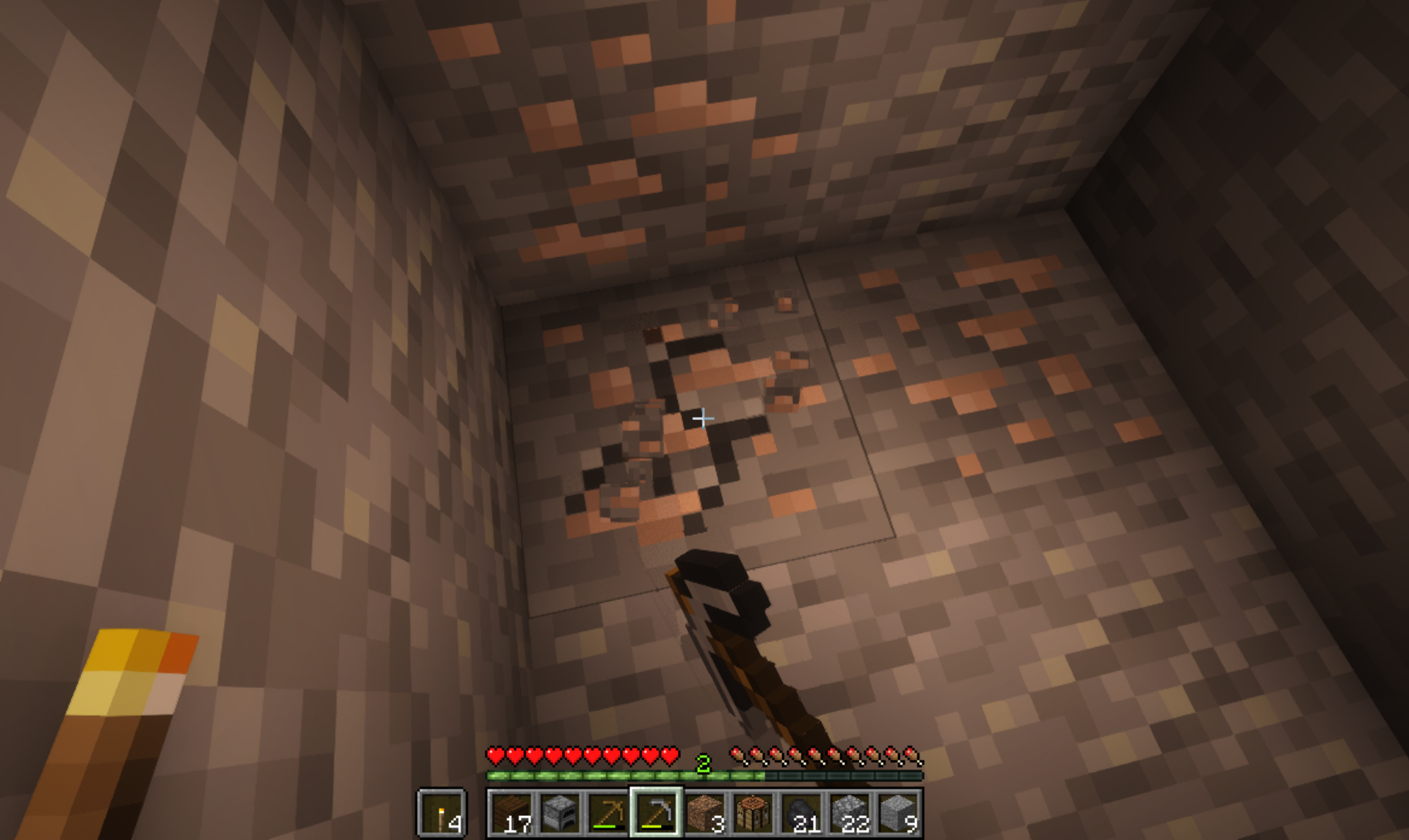}
    }
    \hfill
    \subfloat[Smelt $2$ iron ingots\label{example:fig9}]{%
      \includegraphics[width=0.3\textwidth]{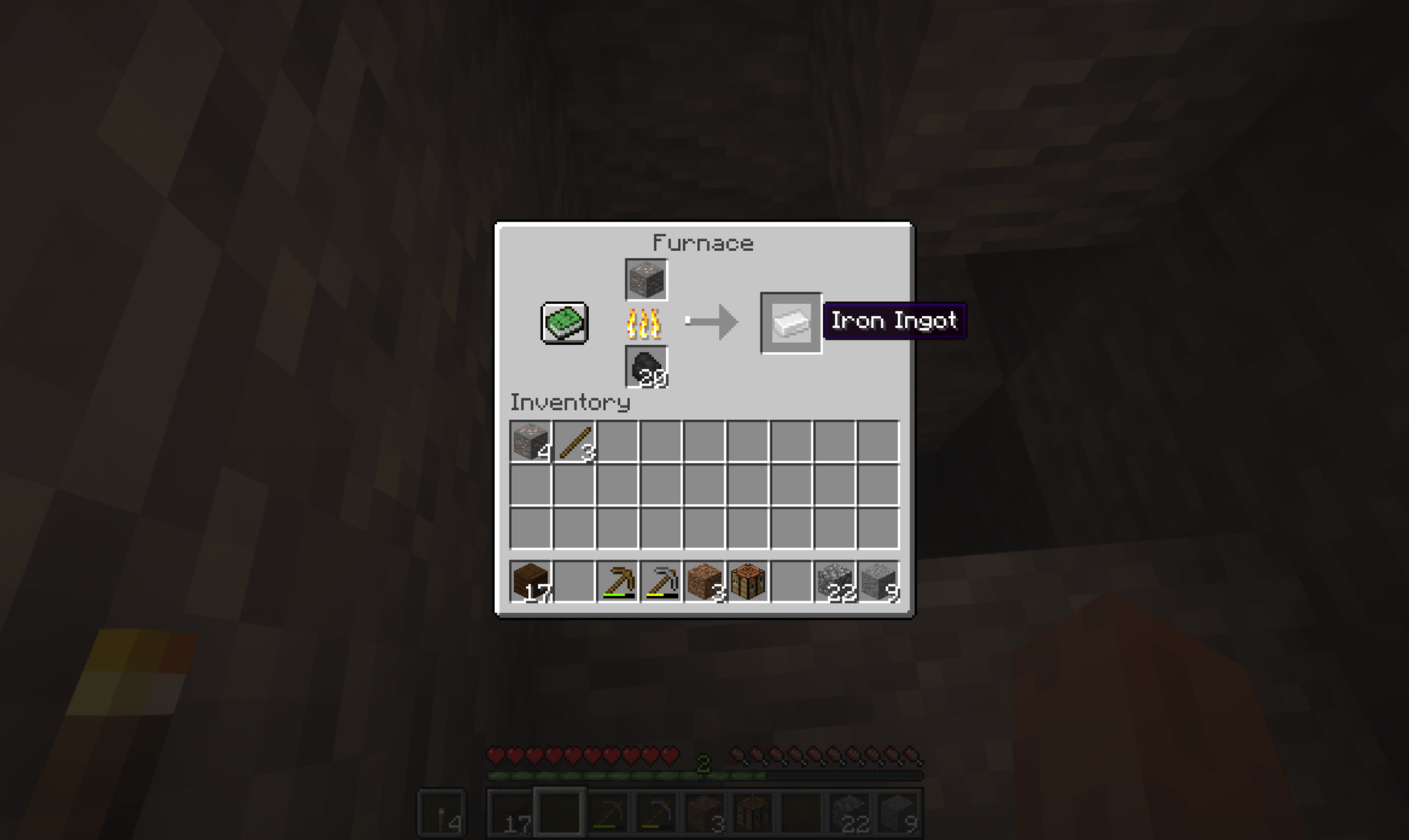}
    }
    \hfill
    \subfloat[Craft $1$ iron sword\label{example:fig10}]{%
      \includegraphics[width=0.3\textwidth]{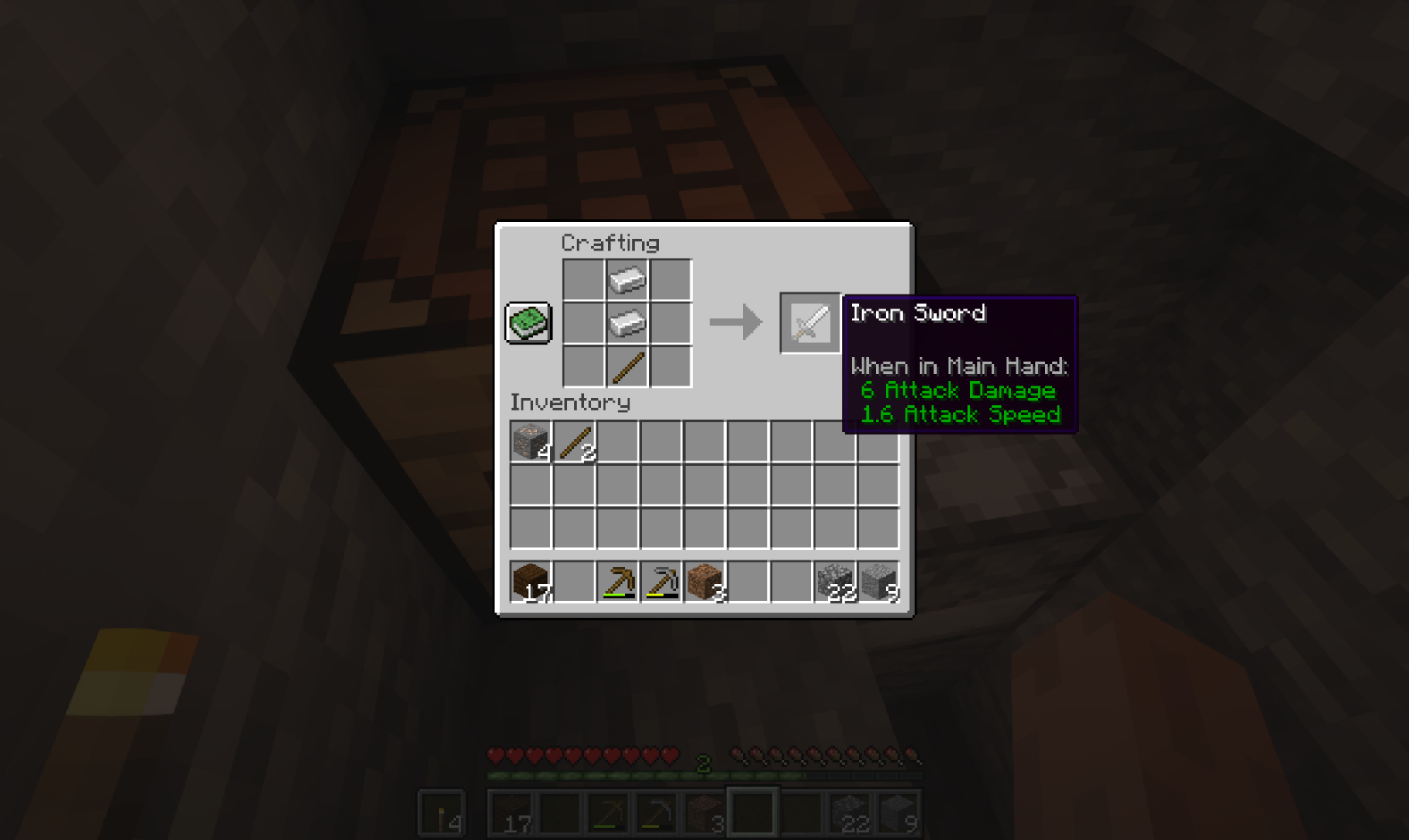}
    }
    \caption{An example of \textit{long-horizon task} ``crafting an iron sword''. The agent must sequentially complete each atomic task in order to successfully craft the iron sword. Failure in any of the atomic tasks will result in the failure of the entire long-horizon task.}
    \label{fig:example_long_horizon}
\end{figure*}

%% file: Supplementary/table/tb_hyperparameter.tex
\begin{table}[ht]
\centering
\caption{Hyperparameter setting for pre-training and finetuning.}
\label{tb:hyper}
\renewcommand\arraystretch{1.2}
\begin{tabular}{ccc}
\toprule[1.2pt]
Hyperparameter & Pre-training & Fine-tuning \\ \hline
Optimizer      & AdamW        & AdamW       \\
Learning Rate  & 0.0001      & 0.00004     \\
Warmup Steps   & 0            & 0           \\
Epochs         & 5            & 10           \\
Batch Size      & 32        & 2048        \\
Num. Frames     & 5M        & 80M         \\
LoRA\_r        & 64           & 64          \\
LoRA\_alpha    & 128          & 128         \\ \bottomrule[1.2pt]
\end{tabular}
\end{table}

%% file: Supplementary/open_example.tex
\begin{table*}[ht]\small%
    \caption{Open-ended instruction examples of “Craft a torch”}
    \label{tab:ins_craft_torch}
    \renewcommand{\arraystretch}{1.5}
    \centering
    \normalsize
    \begin{tabular}{p{1cm}p{12cm}}
        \hline
         \textbf{Planner}& \textbf{Instruction}\\
         \hline
        \multirow{11}{*}{\parbox{6cm}{\textbf{GPT4V}}} &\textbf{Plan 1}
1. First, I need to locate and chop down a couple of trees to collect 4 logs.
2. Next, I should craft these logs into 13 planks.
3. Then, I need to use some of the planks to craft 3 sticks.
4. I will use some of the planks to craft a crafting table.
5. Next, I should craft a wooden pickaxe using some of the planks and sticks.
6. With the wooden pickaxe, I need to find and mine at least 1 coal from a nearby coal deposit or cave.
7. Finally, I can use one stick and one coal to craft a torch.
\\ &\textbf{Plan 2}
1. First, I need to find a tree to gather 4 logs.
2. Next, I have to craft the logs into planks.
3. From the planks, I will craft 3 sticks.
4. I need to craft a crafting table using some of the planks.
5. I should then craft a wooden pickaxe to mine coal.
6. I will locate and mine 1 piece of coal using the wooden pickaxe.
7. Finally, I will use the coal and sticks to craft a torch.
\\ &\textbf{Plan 3}
1. I have to find a tree to get 4 logs.
2. Next, I have to craft planks from the logs.
3. Get sticks from some of the planks.
4. I need a crafting table, which requires planks to craft.
5. I should craft a wooden pickaxe using the crafting table.
6. I need to find and mine some coal using the wooden pickaxe.
7. Finally, with the stick and coal, I can craft a torch.
\\
         \hline

\multirow{15}{*}{\parbox{6cm}{\textbf{GLM4V}}} &\textbf{Plan 1}
1. Locate and chop down a tree to gather 4 logs.
2. Convert the logs into 13 planks using the crafting menu.
3. Use the planks to craft a crafting table.
4. Craft 3 sticks from the remaining planks.
5. With the crafting table and sticks, craft a wooden pickaxe if one is not already available.
6. Use the wooden pickaxe to dig for coal or kill some Creepers or zombies for coal drops.
7. Once 1 coal is obtained, use the crafting table to combine it with a stick to craft 1 torch.
\\ &\textbf{Plan 2}
1. Locate and chop down a tree to gather the 4 logs required.
2. Convert the logs into 13 planks using the crafting menu, as this will be the material needed for crafting the crafting table and the wooden pickaxe.
3. Use some of the planks to craft a crafting table, which is necessary for advanced crafting recipes.
4. Craft a wooden pickaxe using some of the planks and sticks. The wooden pickaxe will be used to mine coal.
5. Gather 3 sticks, which can be obtained from the logs or by crafting them from planks.
6. Use the wooden pickaxe to mine coal. Only 1 coal is needed to craft a torch.
7. Once all materials are gathered, open the crafting table and combine 1 coal and 1 stick to craft a torch.
\\ &\textbf{Plan 3}
1. Locate and gather 4 logs from the nearest trees.
2. Convert the logs into 13 planks using the crafting menu.
3. Use some planks to craft a crafting table for more complex recipes.
4. Craft a wooden pickaxe using some planks and sticks to gather coal and other resources.
5. Convert some planks into 3 sticks required for crafting torches and the wooden pickaxe.
6. Use the wooden pickaxe to mine coal ore, ensuring to gather at least 1 coal.
7. With all materials gathered, open the crafting table and craft 1 torch using 1 coal and 1 stick.
\\
         \hline

    \end{tabular}
    \label{tab:example1}
\end{table*}
\begin{table*}[ht]\footnotesize%
    \caption{Open-ended instruction examples of “Craft a rail”}
    \label{tab:ins_craft_rail}
    \renewcommand{\arraystretch}{1.2}
    \centering
    \normalsize
    \begin{tabular}{p{1cm}p{12cm}}
        \hline
         \textbf{Planner}& \textbf{Instruction}\\
         \hline
        \multirow{15}{*}{\parbox{6cm}{\textbf{GPT4V}}} &\textbf{Plan 1}
1. First, locate and punch down a tree to acquire 7 logs.
2. Convert these logs into planks to obtain a total of 25 planks.
3. Use some of these planks to craft 5 sticks.
4. Create a crafting table using 4 planks.
5. Craft a wooden pickaxe using 3 planks and 2 sticks.
6. Use the wooden pickaxe to mine 11 cobblestones.
7. Craft a furnace using 8 cobblestones.
8. Craft a stone pickaxe using 3 cobblestones and 2 sticks.
9. Use the stone pickaxe to mine 6 iron ores.
10. Smelt the iron ores in the furnace to create 6 iron ingots.
11. Finally, craft a rail using the iron ingots and sticks.
\\ &\textbf{Plan 2}
1. I have to find a tree to get 7 logs.
2. Next, I have to craft the logs into 25 planks.
3. Craft 5 sticks from the planks.
4. I need to create a crafting table.
5. Use the crafting table to craft a wooden pickaxe.
6. Use the wooden pickaxe to mine 11 pieces of cobblestone.
7. I'll then craft a furnace using the cobblestone.
8. Craft a stone pickaxe with the cobblestone and sticks.
9. Use the stone pickaxe to mine 6 pieces of iron ore.
10. Smelt the iron ore in the furnace to obtain 6 iron ingots.
11. Finally, use the crafting table to craft a rail using the iron ingots and sticks.
\\ &\textbf{Plan 3}
1. I have to find a tree to get 7 logs.
2. Next, I need to craft planks from the logs.
3. Get sticks from the planks.
4. Craft a crafting table using some of the planks.
5. Craft a wooden pickaxe with sticks and planks to mine stone.
6. Mine 11 cobblestone with the wooden pickaxe.
7. Craft a furnace using the cobblestone.
8. Craft a stone pickaxe using cobblestone and sticks.
9. Use the stone pickaxe to mine 6 iron ore.
10. Smelt the iron ore in the furnace to get 6 iron ingots.
11. Finally, use the iron ingots and sticks to craft a rail.
\\
         \hline

\multirow{18}{*}{\parbox{6cm}{\textbf{GLM4V}}} &\textbf{Plan 1}
1. Locate and gather 7 logs from nearby trees.
2. Use the crafting menu to convert the logs into 25 planks.
3. Craft 5 sticks from the planks.
4. Place down a crafting table to access more crafting recipes.
5. Use the crafting table to craft a wooden pickaxe.
6. With the wooden pickaxe, mine 11 cobblestones from the ground or from stone blocks.
7. Craft a furnace using the cobblestones.
8. Upgrade the wooden pickaxe to a stone pickaxe by crafting it at the crafting table with the cobblestones.
9. Use the stone pickaxe to mine 6 iron ore from the appropriate rock layers.
10. Smelt the 6 iron ore in the furnace to produce 6 iron ingots.
11. Finally, use the crafting table to craft 1 rail using the 6 iron ingots and 1 stick.
\\ &\textbf{Plan 2}
1. Locate and gather 7 logs from nearby trees.
2. Use the crafting menu to convert the logs into 25 planks.
3. Craft 5 sticks using the planks in the crafting menu.
4. Place a crafting table to enable the crafting of more complex items.
5. Craft a wooden pickaxe using some of the planks and sticks.
6. Use the wooden pickaxe to mine 11 cobblestones from stone outcrops.
7. Construct a furnace using the cobblestones.
8. Upgrade to a stone pickaxe by crafting it with the cobblestones and sticks.
9. Use the stone pickaxe to mine 6 iron ore from iron veins.
10. Smelt the 6 iron ore in the furnace to produce 6 iron ingots.
11. Finally, use the crafting table to craft 1 rail using the iron ingots and sticks.
\\ &\textbf{Plan 3}
1. Find and chop down a tree to gather 7 logs.
2. Use the crafting menu to convert the logs into 25 planks.
3. Craft 5 sticks from the planks.
4. Place down a crafting table to access more crafting options.
5. Craft a wooden pickaxe using some of the planks and sticks.
6. Use the wooden pickaxe to mine 11 cobblestones from stone.
7. Construct a furnace using some of the cobblestones.
8. Upgrade the wooden pickaxe to a stone pickaxe by crafting it with the cobblestones.
9. Search for and mine 6 iron ore blocks with the stone pickaxe.
10. Smelt the iron ore in the furnace to produce 6 iron ingots.
11. Finally, use the crafting table to craft 1 rail using the iron ingots and sticks.
\\
         \hline

    \end{tabular}
    \label{tab:example2}
\end{table*}
\begin{table*}[ht]\footnotesize%
    \caption{Open-ended instruction examples of “Craft a golden shovel”}
    \label{tab:ins_craft_golden_shovel}
    \renewcommand{\arraystretch}{1.1}
    \centering
    \normalsize
    \begin{tabular}{p{1cm}p{15cm}}
        \hline
         \textbf{Planner}& \textbf{Instruction}\\
         \hline
        \multirow{18}{*}{\parbox{6cm}{\textbf{GPT4V}}} &\textbf{Plan 1}
1. Start by finding and chopping down trees to collect 9 logs.
2. Convert the logs into 35 planks using the crafting interface.
3. Use some of the planks to craft 8 sticks.
4. Create a crafting table using some of the planks.
5. Craft a wooden pickaxe.
6. Mine 19 cobblestone.
7. Craft a furnace using 8 of the cobblestones.
8. Craft a stone pickaxe.
9. Mine 3 iron ores using the stone pickaxe.
10. Smelt the iron ore in the furnace.
11. Craft an iron pickaxe using sticks and iron ingots.
12. Locate and mine a gold block using the iron pickaxe.
13. Smelt the gold block in the furnace to get 1 gold ingot.
14. Craft a golden shovel using the gold ingot and sticks.
\\ &\textbf{Plan 2}
1. Start by finding and chopping down trees to collect 9 logs.
2. Use the logs to craft planks.
3. Craft sticks.
4. Use some of the planks to craft a crafting table.
5. Use the crafting table to craft a wooden pickaxe.
6. Use the wooden pickaxe to mine 19 pieces of cobblestone.
7. Use some of the cobblestone to craft a furnace.
8. Use the remaining cobblestone to craft a stone pickaxe.
9. Mine iron ore with the stone pickaxe.
10. Smelt the iron ore in the furnace to obtain 3 iron ingots.
11. Use 2 of the iron ingots and sticks to craft an iron pickaxe.
12. Use the iron pickaxe to mine gold ore or find a gold block; you need at least 1 gold ingot.
13. If mining gold ore, smelt it in the furnace to obtain gold ingots.
14. Use the gold ingot and stick to craft a golden shovel.
\\ &\textbf{Plan 3}
1. Find and cut down trees to collect 9 logs.
2. Use the logs to craft 35 planks.
3. Craft 8 sticks using the planks.
4. Use some of the planks to craft a crafting table.
5. Craft a wooden pickaxe using planks and sticks.
6. Mine cobblestone until you have 19 pieces.
7. Use some cobblestone and the crafting table to craft a furnace.
8. Craft a stone pickaxe using cobblestone and remaining sticks.
9. Use the stone pickaxe to mine 3 pieces of iron ore.
10. Smelt the iron ore in the furnace to get 3 iron ingots.
11. Craft an iron pickaxe.
12. Use the iron pickaxe to mine gold ore to obtain gold blocks.
13. Use 1 gold ingot to craft a golden shovel. 
14. use a gold ingot and sticks to craft a golden shovel using the crafting table.
\\
         \hline

\multirow{21}{*}{\parbox{6cm}{\textbf{GLM4V}}} &\textbf{Plan 1}
1. Find and chop down trees to gather 9 logs.
2. Convert the logs into 35 planks using the crafting menu.
3. Craft 8 sticks from the planks.
4. Build a crafting table to expand the crafting options.
5. Craft a wooden pickaxe with the planks and sticks.
6. Cine cobblestone until you have a total of 19.
7. Craft a furnace using 8 cobblestone.
8. Use the furnace to smelt the cobblestone into 19 stone, then use the crafting table to craft a stone pickaxe with the stone and sticks.
9. Use the stone pickaxe to mine 3 iron ore.
10. Smelt the iron ore in the furnace to produce 3 iron ingots.
11. Craft an iron pickaxe using the iron ingots and sticks at the crafting table.
12. Use the iron pickaxe to mine a gold block.
13. Smelt the gold block in the furnace to get 1 gold ingot.
14. Finally, use the crafting table to craft the golden shovel using 1 gold ingot and 2 sticks.
\\ &\textbf{Plan 2}
1. Find and gather 9 logs from nearby trees.
2. Convert the logs into 35 planks using a crafting table.
3. Craft 8 sticks from the planks.
4. Build a crafting table.
5. Use the crafting table to craft a wooden pickaxe with the planks and sticks.
6. Use the wooden pickaxe to mine 19 cobblestones from stone.
7. Construct a furnace to smelt materials.
8. Craft a stone pickaxe using the cobblestones and sticks.
9. Use the stone pickaxe to mine 3 iron ore from iron deposits.
10. Smelt the 3 iron ore in the furnace to produce 3 iron ingots.
11. Craft an iron pickaxe with the iron ingots and sticks.
12. Use the iron pickaxe to mine a gold block.
13. Smelt the gold block in the furnace to get 1 gold ingot.
14. Finally, craft the golden shovel using the gold ingot and stick in the crafting table.
\\ &\textbf{Plan 3}
1. Find and gather 9 logs from nearby trees.
2. Use the crafting menu to convert the logs into 35 planks.
3. Craft 8 sticks from the planks.
4. Craft a crafting table using some of the planks.
5. Use the crafting table to craft a wooden pickaxe with some of the sticks and planks.
6. Use the wooden pickaxe to mine 19 cobblestones from stone.
7. Craft a furnace using some of the cobblestones.
8. Use the furnace to smelt the cobblestones into 19 stone, then craft a stone pickaxe with the sticks and stone.
9. Mine 3 iron ore from the appropriate rock faces.
10. Smelt the iron ore in the furnace to produce 3 iron ingots.
11. Use the crafting table to craft an iron pickaxe with the iron ingots and sticks.
12. Use the iron pickaxe to mine a gold block.
13. Smelt the gold block in the furnace to produce 1 gold ingot.
14. Finally, use the crafting table to craft the golden shovel with the gold ingot and 2 sticks. 
\\
         \hline

    \end{tabular}
    \label{tab:example3}
\end{table*}
\begin{table*}[ht]\footnotesize%
    \caption{Open-ended instruction examples of “Craft a diamond pickaxe”}
    \label{tab:ins_craft_diamond_pickaxe}
    \renewcommand{\arraystretch}{1.2}
    \centering
    \normalsize
    \begin{tabular}{p{1cm}p{15cm}}
        \hline
         \textbf{Planner}& \textbf{Instruction}\\
         \hline
        \multirow{18}{*}{\parbox{6cm}{\textbf{GPT4V}}} &\textbf{Plan 1}
1. First, locate and chop down trees to gather 9 logs.
2. Convert the logs into 35 planks.
3. Use some of the planks to craft 8 sticks.
4. Create a crafting table using 4 planks.
5. With the crafting table, craft a wooden pickaxe using planks and sticks.
6. Use the wooden pickaxe to collect at least 11 cobblestones from stone.
7. With the cobblestones, craft a furnace.
8. Use more cobblestones to craft a stone pickaxe.
9. Use the stone pickaxe to mine 3 iron ores.
10. Smelt the iron ores into 3 iron ingots using the furnace.
11. Craft an iron pickaxe with iron ingots and sticks.
12. Use the iron pickaxe to mine 3 diamonds.
13. Finally, craft the diamond pickaxe using the diamonds and sticks.
\\ &\textbf{Plan 2}
1. Begin by locating a tree to gather 9 logs.
2. Convert the logs into planks.
3. Use some of the planks to craft sticks.
4. Craft a crafting table to facilitate further crafting.
5. Create a wooden pickaxe to start mining stone.
6. Mine 11 cobblestones.
7. Use the cobblestones to craft a furnace.
8. Craft a stone pickaxe to advance to mining iron ore.
9. Locate and mine 3 iron ores using the stone pickaxe.
10. Smelt the iron ores in the furnace to produce 3 iron ingots.
11. Use two of the iron ingots and sticks to craft an iron pickaxe.
12. Locate a diamond ore vein and mine 3 diamonds using the iron pickaxe.
13. Finally, craft the diamond pickaxe with the 3 diamonds and sticks.
\\ &\textbf{Plan 3}
1. I need to find a tree to gather 9 logs.
2. Convert the logs into 35 planks.
3. Use some planks to craft 8 sticks.
4. Craft a crafting table using some of the planks.
5. With the crafting table, make a wooden pickaxe to mine stone.
6. Mine 11 cobblestones using the wooden pickaxe.
7. Craft a furnace using the collected cobblestones.
8. Make a stone pickaxe to efficiently mine iron ore.
9. Find and mine 3 iron ores using the stone pickaxe.
10. Smelt the iron ores in the furnace to produce 3 iron ingots.
11. Craft an iron pickaxe using some planks, sticks, and the iron ingots.
12. Locate a source of diamonds and use the iron pickaxe to mine 3 diamonds.
13. Finally, use the crafting table along with the sticks and diamonds to craft a diamond pickaxe.
\\
         \hline

\multirow{20}{*}{\parbox{6cm}{\textbf{GLM4V}}} &\textbf{Plan 1}
1. Start by finding trees to gather 9 logs.
2. Convert the logs into 35 planks.
3. Craft 8 sticks from the planks.
4. Build a crafting table to expand crafting options.
5. Use the crafting table to make a wooden pickaxe.
6. With the wooden pickaxe, mine 11 cobblestones.
7. Craft a stone pickaxe using the cobblestones.
8. Construct a furnace to smelt materials.
9. Use the stone pickaxe to mine 3 iron ores.
10. Smelt the iron ores in the furnace to produce 3 iron ingots.
11. Craft an iron pickaxe with the iron ingots.
12. Locate and mine 3 diamonds, which may require exploring caves or mining deep underground with the iron pickaxe.
13. Finally, use the crafting table to combine the diamonds and the sticks to craft a diamond pickaxe.
\\ &\textbf{Plan 2}
1. Locate and chop down trees to gather a total of 9 logs.
2. Use the logs to craft 35 planks, as this will be the base material for several other items.
3. Craft 8 sticks from the planks, which will be needed to create various tools.
4. Craft a crafting table using some of the planks, as this is essential for crafting more complex items.
5. Use the crafting table to craft a wooden pickaxe with the planks and sticks, which will be used to mine cobblestone.
6. Mine 11 cobblestone using the wooden pickaxe.
7. Craft a stone pickaxe with the cobblestone and sticks, as this is more durable and efficient for mining.
8. Construct a furnace using the cobblestone, which will be required to smelt ores into ingots.
9. Mine 3 iron ore with the stone pickaxe.
10. Smelt the iron ore in the furnace to produce 3 iron ingots.
11. Craft an iron pickaxe using the iron ingots and sticks, as this is necessary to mine diamonds.
12. Find and mine diamond ore with the iron pickaxe to gather 3 diamonds.
13. Finally, use the crafting table to craft a diamond pickaxe with the diamonds and sticks, completing the task.
\\ &\textbf{Plan 3}
1. Begin by locating trees to gather 9 logs.
2. Convert the logs into 35 planks.
3. Craft 8 sticks from the planks.
4. Build a crafting table to expand crafting options.
5. Use the crafting table to craft a wooden pickaxe.
6. With the wooden pickaxe, mine 11 cobblestones.
7. Craft a stone pickaxe using the cobblestones.
8. Construct a furnace to smelt materials.
9. Use the stone pickaxe to mine 3 iron ore.
10. Smelt the iron ore in the furnace to produce 3 iron ingots.
11. Craft an iron pickaxe with the iron ingots.
12. Locate and mine diamond ore with the iron pickaxe to obtain 3 diamonds.
13. Finally, use the crafting table to craft a diamond pickaxe with the diamonds.
\\
         \hline

    \end{tabular}
    \label{tab:example4}
\end{table*}
\begin{table*}[ht]\footnotesize%
    \caption{Open-ended instruction examples of “Craft a compass”}
    \label{tab:ins_craft_compass}
    \renewcommand{\arraystretch}{1.2}
    \centering
    \normalsize
    \begin{tabular}{p{1cm}p{15cm}}
        \hline
         \textbf{Planner}& \textbf{Instruction}\\
         \hline
        \multirow{18}{*}{\parbox{8cm}{\textbf{GPT4V}}} &\textbf{Plan 1}
1. Locate a tree and gather 10 logs.
2. Convert the logs into 38 planks.
3. Craft 8 sticks from the planks.
4. Use some planks to craft a crafting table.
5. With the crafting table, craft a wooden pickaxe.
6. Mine cobblestone until you have 12 pieces.
7. Upgrade your tools by crafting a stone pickaxe.
8. Construct a furnace using some of the cobblestone.
9. Use the stone pickaxe to mine 7 iron ore.
10. Smelt the iron ore in the furnace to produce 7 iron ingots.
11. Craft an iron pickaxe using the iron ingots.
12. Find and mine redstone using the iron pickaxe to obtain 1 redstone dust.
13. Finally, use the remaining iron ingots and redstone dust at the crafting table to craft a compass.
\\ &\textbf{Plan 2}
1. Start by finding and chopping down trees to collect 10 logs.
2. Use the logs to craft 38 planks.
3. Craft sticks using the planks to get 8 sticks.
4. Craft a crafting table from the planks.
5. Use the crafting table to create a wooden pickaxe.
6. Locate and mine 12 pieces of cobblestone using the wooden pickaxe.
7. Craft a stone pickaxe from cobblestone and sticks.
8. Use the cobblestone to craft a furnace.
9. Mine 7 iron ores using the stone pickaxe.
10. Smelt the iron ore in the furnace to get 7 iron ingots.
11. Craft an iron pickaxe using iron ingots and sticks.
12. Search for and mine redstone using the iron pickaxe.
13. Finally, use the iron ingots and redstone to craft a compass.
\\ &\textbf{Plan 3}
1. First, I need to find a tree and collect 10 logs.
2. Convert the logs into planks, ensuring I get at least 38 planks.
3. Use some of the planks to craft sticks, collecting at least 8 sticks.
4. Use some of the planks to create a crafting table.
5. Craft a wooden pickaxe.
6. Locate cobblestone and mine at least 12 pieces using the wooden pickaxe.
7. Use some cobblestone to craft a furnace.
8. Craft a stone pickaxe for improved mining.
9. Search for iron ore and mine at least 7 using the stone pickaxe.
10. Smelt the iron ore in the furnace to produce 7 iron ingots.
11. Craft an iron pickaxe using some of the iron ingots and sticks.
12. Locate a redstone ore and mine it using the iron pickaxe to obtain redstone.
13. Finally, use 4 of the iron ingots and the redstone to craft a compass at the crafting table.
\\
         \hline

\multirow{18}{*}{\parbox{8cm}{\textbf{GLM4V}}} &\textbf{Plan 1}
1. Locate and gather 10 logs from nearby trees.
2. Convert the logs into 38 planks using the crafting menu.
3. Craft 8 sticks from the planks.
4. Build a crafting table to access more crafting options.
5. Use the crafting table to make a wooden pickaxe.
6. With the wooden pickaxe, mine 12 cobblestones from stone.
7. Craft a stone pickaxe using the cobblestones and the crafting table.
8. Construct a furnace to smelt materials.
9. Use the stone pickaxe to mine 7 iron ores from the appropriate rock formations.
10. Smelt the 7 iron ores in the furnace to produce 7 iron ingots.
11. Craft an iron pickaxe with the iron ingots at the crafting table.
12. Mine for redstone, which can sometimes be found in the ground.
13. Use the crafting table to craft a compass with the iron ingot and redstone.
\\ &\textbf{Plan 2}
1. Locate and chop down trees to gather the required 10 logs.
2. Convert the logs into 38 planks using the crafting menu.
3. Craft 8 sticks from the planks.
4. Use some planks to craft a crafting table.
5. With the crafting table and the sticks and planks crafted, make a wooden pickaxe.
6. Use the wooden pickaxe to mine 12 cobblestone.
7. Craft a stone pickaxe using the cobblestone.
8. Construct a furnace with the cobblestone.
9. Use the stone pickaxe to mine 7 iron ore.
10. Smelt the iron ore in the furnace to produce 7 iron ingots.
11. Craft an iron pickaxe using the iron ingots and the crafting table.
12. Search for redstone; gather at least 1 redstone.
13. Finally, use the crafting table to combine the redstone with the iron ingots to craft a compass.
\\ &\textbf{Plan 3}
1. Find and chop down trees to gather 10 logs.
2. Convert the logs into 38 planks.
3. Craft 8 sticks from the planks.
4. Craft a crafting table using some of the planks.
5. Use the crafting table to craft a wooden pickaxe.
6. With the wooden pickaxe, mine cobblestone until you have at least 12.
7. Upgrade the wooden pickaxe to a stone pickaxe using the cobblestone.
8. Construct a furnace out of the cobblestone.
9. Use the stone pickaxe to mine 7 iron ore.
10. Smelt the 7 iron ore in the furnace to produce 7 iron ingots.
11. Craft an iron pickaxe using the iron ingots and sticks at the crafting table.
12. Search for redstone; you will need at least 1 piece.
13. Finally, use the crafting table to combine the redstone with the iron ingots to craft a compass.
\\
         \hline

    \end{tabular}
    \label{tab:example5}
\end{table*}

%% file: Supplementary/table/optimus1_benchmark_result/wooden_result.tex
\begin{table*}[ht]
\centering
\caption{The results of Optimus-2 on the Wood Group, Stone Group, and Iron Group. SR denotes success rate.}
\label{tab:wooden_result}
\resizebox{0.8\textwidth}{!}{%
\renewcommand\arraystretch{1.1}
\begin{tabular}{llccc}
\toprule[1.5pt]
\textbf{Group}& \textbf{Task} & \textbf{Sub-Goal Num.} & \textbf{SR} & \textbf{Eval Times} \\ \hline

\multirow{10}{*}{Wood} & Craft a wooden shovel  & 6 & 100.00   & 40 \\
&Craft a wooden pickaxe & 5 & 100.00 & 30 \\
&Craft a wooden axe     & 5 & 97.37  & 38 \\
&Craft a wooden hoe     & 5 & 100.00  & 30 \\
&Craft a stick          & 4 & 100  & 30 \\
&Craft a crafting table & 3 & 93.02  & 43 \\
&Craft a wooden sword   & 5 & 100.00  & 30 \\
&Craft a chest          & 4 & 100.00  & 30 \\
&Craft a bowl           & 4 & 100.00  & 30 \\ 
& Craft a ladder           & 4 & 100.00  & 30 \\
\hline

\multirow{9}{*}{Stone}&Craft a stone shovel  & 8  & 89.47 & 57 \\
&Craft a stone pickaxe & 10 & 98.00  & 50 \\
&Craft a stone axe     & 10 & 94.44  & 54 \\
&Craft a stone hoe     & 8  & 95.74  & 47 \\
&Craft a charcoal      & 9  & 85.71  & 42 \\
&Craft a smoker        & 9  & 90.00  & 40 \\
&Craft a stone sword   & 8  & 95.45 & 44 \\
&Craft a furnace       & 9  & 94.44  & 36 \\
&Craft a torch         & 8  & 89.36  & 47 \\
\hline
\multirow{16}{*}{Iron}&Craft an iron shovel   & 13 & 52.08   & 48 \\
&Craft an iron pickaxe  & 13 & 56.00  & 50 \\
&Craft an iron axe      & 13 & 48.15  & 54 \\
&Craft an iron hoe     & 13 & 56.60   & 53 \\
&Craft a bucket         & 13 & 45.10   & 51 \\
&Craft a hopper         & 14 & 54.90  & 51 \\
&Craft a rail           & 13 & 51.02 & 49 \\
&Craft an iron sword    & 12 & 56.52  & 46 \\
&Craft a shears         & 12 & 48.28 & 58 \\
&Craft a smithing table & 12 & 53.33   & 45 \\
&Craft a tripwire hook  & 13 & 55.56   & 45 \\
&Craft a chain          & 13 & 52.17   & 46 \\
&Craft an iron bars     & 12 & 51.06   & 47 \\
&Craft an iron nugget   & 12 & 54.55   & 44 \\
&Craft a blast furnace  & 14 & 52.27   & 44 \\
&Craft a stonecutter    & 13 & 52.27   & 44 \\
\bottomrule[1.5pt]
\end{tabular}%
}
\end{table*}

%% file: Supplementary/table/optimus1_benchmark_result/golden_result.tex
\begin{table*}[ht]
\centering
\caption{The results of Optimus-2 on the Gold group, Diamond Group, Redstone Group, and Armor Group. SR denotes success rate.}
\label{tab:gold_result}
\resizebox{0.8\textwidth}{!}{%
\renewcommand\arraystretch{1.2}
\begin{tabular}{llccc}
\toprule[1.5pt]
\textbf{Group} & \textbf{Task} & \textbf{Sub Goal Num.} & \textbf{SR} &  \textbf{Eval Times} \\ \hline

\multirow{6}{*}{Gold}&Craft a golden shovel         & 16 & 8.93  & 56 \\
&Craft a golden pickaxe        & 16 & 11.29 & 62 \\
&Craft a golden axe            & 16 & 8.93  & 56 \\
&Craft a golden hoe            & 16 & 8.96  & 67 \\
&Craft a golden sword          & 16 & 8.20  & 61 \\
&Smelt and craft an golden ingot & 15 & 9.68 & 62 \\ 
\hline
\multirow{7}{*}{Diamond}&Craft a diamond shovel      & 15 & 15.91 & 44 \\
&Craft a diamond pickaxe     & 15 & 11.76 & 34 \\
&Craft a diamond axe         & 16 & 11.00  & 36 \\
&Craft a diamond hoe         & 15 & 15.91  & 44 \\
&Craft a diamond sword       & 15 & 11.11 & 36 \\
&Dig down and mine a diamond & 15 & 11.42  & 35 \\
&Craft a jukebox             & 15 & 13.15 & 38 \\
\hline
\multirow{6}{*}{Redstone}&Craft a piston          & 16 & 28.33  & 60 \\
&Craft a redstone torch  & 16 & 27.69 & 65 \\
&Craft an activator rail & 18 & 25.81 & 62 \\
&Craft a compass         & 23 & 28.36 & 67 \\
&Craft a dropper         & 16 & 30.30  & 66 \\
&Craft a note block      & 16 & 25.40  & 63 \\ 
\hline
\multirow{13}{*}{Armor}&Craft shield             & 14 & 45.16 & 62 \\
&Craft iron chestplate    & 14 & 43.86 & 57 \\
&Craft iron boots         & 14 & 40.35 & 57 \\
&Craft iron leggings      & 14 & 8.57  & 35 \\
&Craft iron helmet        & 14 & 47.46 & 56 \\
&Craft diamond helmet     & 17 & 9.09  & 33 \\
&Craft diamond chestplate & 17 & 7.89  & 38 \\
&Craft diamond leggings   & 17 & 5.41  & 37 \\
&Craft diamond boots      & 17 & 12.50 & 40 \\
&Craft golden helmet      & 17 & 13.89 & 36 \\
&Craft golden leggings    & 17 & 12.20 & 41 \\
&Craft golden boots       & 17 & 10.26  & 39 \\
&Craft golden chestplate  & 17 & 10.00  & 40 \\
\bottomrule[1.5pt]
\end{tabular}%
}
\end{table*}

%% file: Supplementary/case_study.tex
\begin{figure*}[htbp]
    \centering
    \includegraphics[width=1\textwidth]{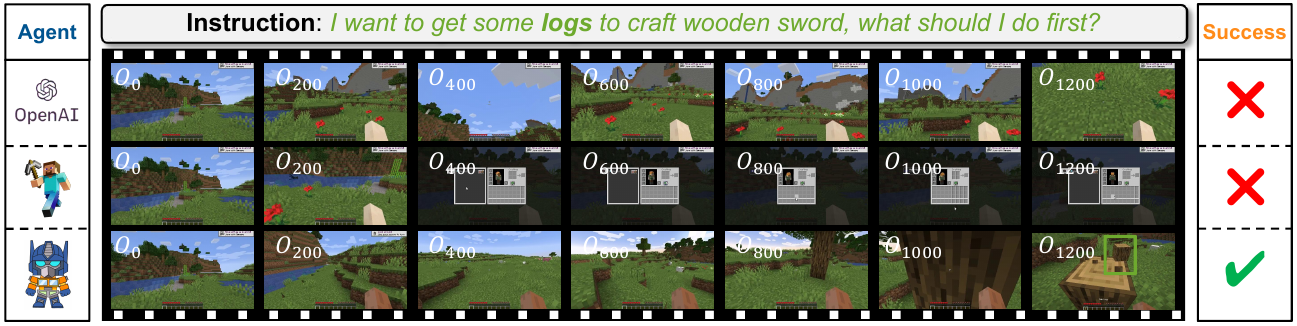}
    \caption{An illustration of VPT (text) \cite{vpt}, STEVE-1 \cite{lifshitz2024steve}, and Optimus-2 executing the open-ended instruction, ``I want to get some logs to craft wooden sword, what should I do first?''. Existing policies are limited by their instruction comprehension abilities and thus fail to complete the task, whereas GOAP leverages the language understanding capabilities of the MLLM, enabling it to accomplish the task.}
    \label{fig:case1}
\end{figure*}

\begin{figure*}[htbp]
    \centering
    \includegraphics[width=1\textwidth]{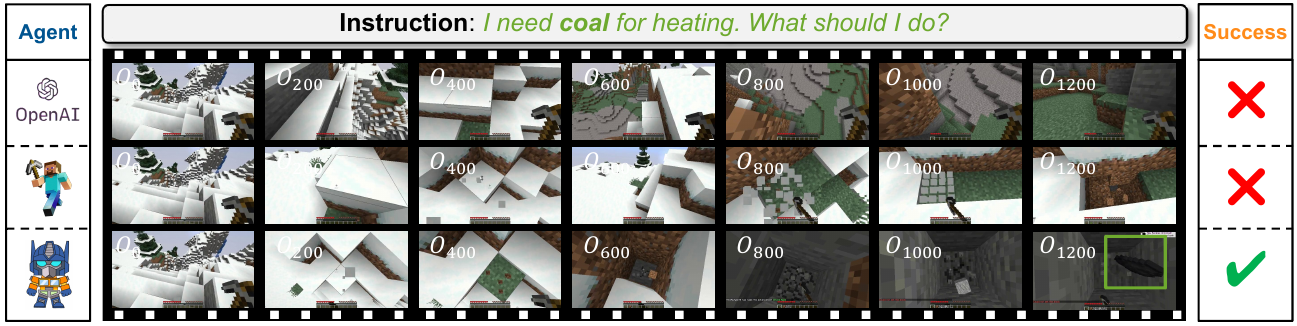}
    \caption{An illustration of VPT (text) \cite{vpt}, STEVE-1 \cite{lifshitz2024steve}, and Optimus-2 executing the open-ended instruction, ``I need coal for heating. What should I do?''. Existing policies are limited by their instruction comprehension abilities and thus fail to complete the task, whereas GOAP leverages the language understanding capabilities of the MLLM, enabling it to accomplish the task.}
    \label{fig:case2}
\end{figure*}

\begin{figure*}[htbp]
    \centering
    \includegraphics[width=1\textwidth]{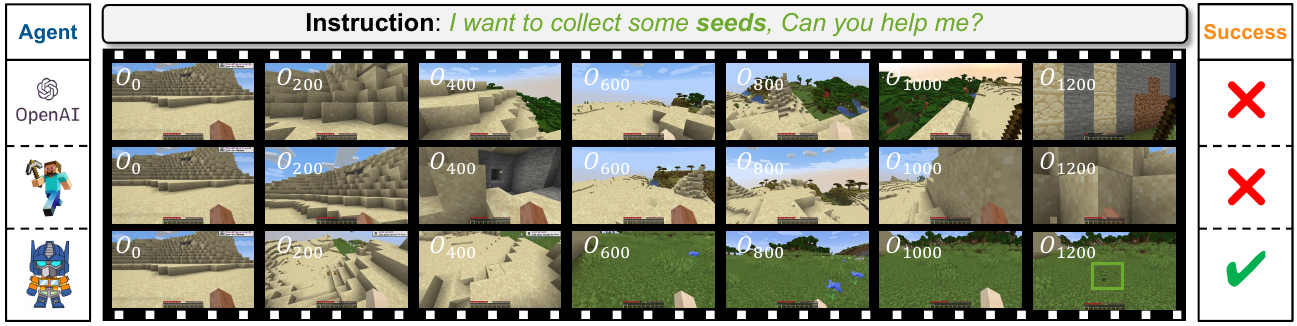}
    \caption{An illustration of VPT (text) \cite{vpt}, STEVE-1 \cite{lifshitz2024steve}, and Optimus-2 executing the open-ended instruction, ``I want to collect some seeds, Can you help me?''. Existing policies are limited by their instruction comprehension abilities and thus fail to complete the task, whereas GOAP leverages the language understanding capabilities of the MLLM, enabling it to accomplish the task.}
    \label{fig:case3}
\end{figure*}